\documentclass[sigconf]{acmart}

\usepackage{multirow}
\usepackage{graphicx}

\copyrightyear{2023}
\acmYear{2023}
\setcopyright{acmlicensed}\acmConference[SA Conference Papers '23]{SIGGRAPH Asia 2023 Conference Papers}{December 12--15, 2023}{Sydney, NSW, Australia}
\acmBooktitle{SIGGRAPH Asia 2023 Conference Papers (SA Conference Papers '23), December 12--15, 2023, Sydney, NSW, Australia}
\acmPrice{15.00}
\acmDOI{10.1145/3610548.3618194}
\acmISBN{979-8-4007-0315-7/23/12}

\acmSubmissionID{538}

\begin{document}

\title{MatFusion: A Generative Diffusion Model for SVBRDF Capture}

\author{Sam Sartor}
\affiliation{%
  \institution{College of William \& Mary}
  \city{Williamsburg}
  \country{USA}
}
\orcid{0009-0001-1915-6887}
\email{slsartor@wm.edu}

\author{Pieter Peers}
\affiliation{%
  \institution{College of William \& Mary}
  \city{Williamsburg}
  \country{USA}
}
\orcid{0000-0001-7621-9808}
\email{ppeers@siggraph.org}

\renewcommand\shortauthors{Sartor and Peers}

\def\equationautorefname~#1\null{Equation~(#1)\null}
\def\sectionautorefname{Section}
\def\subsectionautorefname{Section}

\def\etal{{et al.}}
\def\ie{{i.e.}}
\def\eg{{e.g.}}

\def\BF{\textbf}
\def\UL{\underline}

\newcommand{\Variant}[1]{\textsc{#1}}

\newlength\rsepwidth
\setlength{\rsepwidth}{0.07em}
\newlength\reswidth
\newlength\resdwidth
\newlength\resinwidth
\newlength\resinoffset
\newcommand{\updatewidths}{
\def\rsep{\hspace{\rsepwidth}}
\setlength{\resdwidth}{2\reswidth}
\addtolength{\resdwidth}{\rsepwidth}
}

\definecolor{srcpolyhaven}{rgb}{0.201062983818605, 0.2764395652150914, 0.7545625726213753}
\definecolor{srcjamie}{rgb}{0.4976548130982128, 0.7850809318550582, 0.23118028042347583}
\definecolor{srcambient}{rgb}{0.4453788599067607, 0.2594613505197329, 0.7545625726213753}
\definecolor{srclookahead}{rgb}{0.8162729714820274, 0.190368116509061, 0.2721389334031778}

\newsavebox\MyPicture

\newlength\tssepwidth
\setlength{\tssepwidth}{2pt}

\NewDocumentCommand{\testsetitembox}%
    {mm}{%
    \savebox\MyPicture{%
        \includegraphics[width=\reswidth]{#1}\!
        \includegraphics[width=\reswidth]{#2}%
    }%
}
\NewDocumentCommand{\testsetitem}%
    {mmm}{%
    \testsetitembox{#2}{#3}
    \begin{tikzpicture}%
    \draw [path picture={%
        \node at (path picture bounding box.center) {%
        \usebox\MyPicture};},draw=#1,line width=2pt]
        (-\tssepwidth,-\tssepwidth)  rectangle (\wd\MyPicture+\tssepwidth,\ht\MyPicture+\tssepwidth);
   \end{tikzpicture}%
}

\providecommand{\DIFdel}[1]{}
\renewcommand{\DIFdel}[1]{}

\providecommand{\DIFadd}[1]{}
\renewcommand{\DIFadd}[1]{\textcolor{blue}{#1}}

\begin{abstract}
  We formulate SVBRDF estimation from photographs as a diffusion task. To
  model the distribution of spatially varying materials, we first train a
  novel unconditional SVBRDF diffusion backbone model on a large set of
  $312,\!165$ synthetic spatially varying material exemplars.  This SVBRDF
  diffusion backbone model, named MatFusion, can then serve as a basis for
  refining a conditional diffusion model to estimate the material properties
  from a photograph under controlled or uncontrolled lighting. Our backbone
  MatFusion model is trained using only a loss on the reflectance properties,
  and therefore refinement can be paired with more expensive rendering methods
  without the need for backpropagation during training.  Because the
  conditional SVBRDF diffusion models are generative, we can synthesize
  multiple SVBRDF estimates from the same input photograph from which the user
  can select the one that best matches the users' expectation.  We demonstrate
  the flexibility of our method by refining different SVBRDF diffusion models
  conditioned on different types of incident lighting, and show that for a
  single photograph under colocated flash lighting our method achieves equal
  or better accuracy than existing SVBRDF estimation methods.
\end{abstract}



\begin{teaserfigure}
  \centering
  \includegraphics[width=\textwidth]{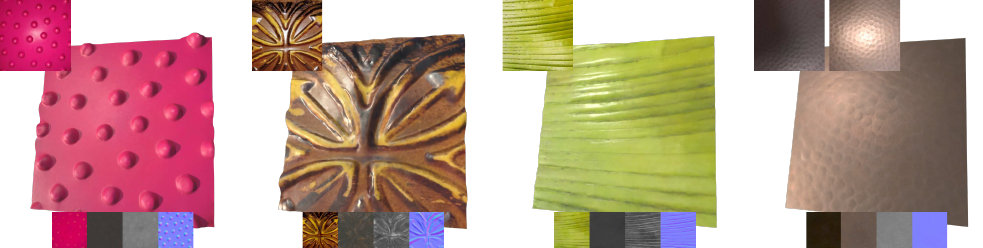}
  \caption{SVBRDF diffusion estimates visualized with integrated normal maps
    and global illumination for four different spatially varying materials
    captured by: a colocated flash photograph (1st and 2nd), a photograph
    captured under uncontrolled natural lighting (3rd), and a flash/no-flash
    image pair (4th).}
  \label{fig:teaser}
\end{teaserfigure}

\maketitle
\def\aw{a}
\def\bw{b}

\def\x{x}
\def\p{p}
\def\pdata{\p_\text{data}}
\def\T{T}
\def\t{t}
\def\N{\mathcal{N}}
\def\s{\sigma}
\def\param{\theta}
\def\D{D_\param}
\def\n{n}
\def\E{\mathbb{E}}
\def\d{\mathbf{d}}
\def\y{y}
\def\c{c}

\section{Introduction}
\label{sec:intro}

Reproducing the visual appearance of real-world spatially varying materials is
a challenging research problem that requires balancing multiple competing
goals such as ease of capture, robustness, accuracy of the reproduction, and
suitability for post-production editing.  The most promising recent solutions
leverage machine learning to produce Spatially Varying Bidirectional
Reflectance Distribution Function (SVBRDF) parameter maps that correspond to
one or more photographs of the target material.  These methods are convenient
and can produce plausible SVBRDFs.  However, SVBRDF modeling is inherently
ambiguous as multiple parameter combinations can explain the
(underconstrained) appearance observations of the material, and there is no
recourse when the inferred property maps fail to reproduce plausible material
properties; there are typically no additional hyper-parameters that can be
tuned to produce alternative solutions.  Furthermore, these machine learning
based methods are trained for a specific type of incident lighting, and
modifying the input lighting often requires a lengthy retraining step and an
appropriate corresponding loss.

Inspired by recent successes in using diffusion
models~\cite{Song:2021:SBG,Karras:2022:EDS,Rombach:2022:HRI} for image
synthesis tasks such as image
restoration~\cite{Dhariwal:2021:DMB,Ho:2020:DDP,Ho:2022:CDM},
super-resolution~\cite{Kadhodaie:2021:SSL,Saharia:2023:ISR}, and
image-to-image translation~\cite{Sasaki:2021:UDU,Saharia:2022:PII} we
formulate SVBRDF estimation as a diffusion task.  Existing diffusion based
image processing methods rely on pre-trained large scale image diffusion
models to sample the distribution of natural images.  However, the
distribution of SVBRDFs differs significantly from natural images.  We
therefore introduce a novel generative diffusion model geared towards
spatially varying materials. We introduce an unconditional backbone diffusion
model, named MatFusion, that synthesizes SVBRDF parameter maps (\ie, diffuse
and specular albedo, specular roughness, and normals).  We leverage ConvNeXt
blocks~\cite{Liu:2022:ACN} instead of the typical Residual
blocks~\cite{He:2016:DRL} commonly used in diffusion models to increase the
number of activations without increasing the parameter count to better model
the $10$ SVBRDF channels (versus $3$ for images).  Furthermore, training
diffusion models typically requires a significantly larger training set than
conventional convolutional neural networks. To support training an SVBRDF
diffusion model, we supplement the INRIA synthetic SVBRDF
dataset~\cite{Deschaintre:2018:SIS} with a new training set constructed from
$1,\!877$ synthetic SVBRDFs, that after augmentation with a novel mixing
strategy, together with the INRIA dataset, grows to $312,\!165$ unique
training exemplars.  Building on the MatFusion backbone, we also introduce
\emph{three} conditional refinements that differ in their input: the classic
colocated camera-flash image, a photograph under uncontrolled natural
lighting, and a flash/no-flash image pair (\autoref{fig:teaser}).  By changing
the seed, all three models can produce a variety of candidate SVBRDF
replicates, from which the SVBRDF that best matches the user's expectation can
be selected.  Our backbone diffusion network is trained using only SVBRDF
parameter losses (\ie, without a rendering loss), and thus no backpropagation
through a differentiable renderer is needed.  This allows us to train the
conditional diffusion network on input images that contain a more complete
characterization of the surface reflectance by integrating the normal maps and
accounting for indirect lighting within the material.  While such indirect
lighting does not contribute significantly for backscatter surface
reflectance, it does impact the visual appearance significantly for more
complex lighting conditions (such as natural lighting).

We demonstrate the efficacy of finetuning the MatFusion backbone and show that
the conditional diffusion networks produce plausible SVBRDFs, and in case of
colocated flash lighting, with equal or better quality than existing methods.

In summary, our contributions are:
\begin{enumerate}
\item MatFusion: a backbone k-diffusion model that generates $10$ channels of reflectance
  properties;
\item three conditional SVBRDF diffusion models refined from the MatFusion
  backbone using a novel direct conditioning strategy; and
\item a training set of $312,\!165$ unique synthetic SVBRDFs.
\end{enumerate}

\section{Related Work}
\label{sec:related}

We focus the discussion of related work on learning-based generative and
inference methods for modeling SVBRDFs.

\paragraph{Direct Inference Methods}
Estimating spatially varying material parameters from a single photograph is a
difficult problem.  Leveraging advances in neural networks,
Li~\etal~\shortcite{Li:2017:MSA} and Ye~\etal~\shortcite{Ye:2018:SIS}
demonstrate plausible SVBRDF capture from a single photograph under unknown
natural lighting, albeit restricted to a predetermined class of materials
(\eg, metals, plastics, etc.)
Deschainte~\etal~\shortcite{Deschaintre:2018:SIS} introduced the de-facto
standard training set of approximately $200,\!000$ synthesized SVBRDFs to
train an inference network, using a novel render loss, that estimates the
SVBRDF property maps from a single photograph lit by a colocated flash
light. Subsequent work further improved the inference accuracy by exploring
novel architectures and loss
functions~\cite{Li:2018:MMS,Sang:2020:SSN,Zhou:2021:ASI,Vecchio:2021:SAS,Guo:2021:HAT}
or supporting multiple input
photographs~\cite{Deschaintre:2019:FSC,Ye:2021:DRS}. Martin~\etal~\shortcite{Martin:2022:MSI}
capture SVBRDFs, albeit without specular albedo, from outdoor photographs that
include ambient occlusion effects.  All of the above methods are trained for a
specific input lighting condition; it is unclear to what degree the
architecture and loss are tuned to the expected lighting, and significantly
changing the lighting condition during capture would require retraining the
network from scratch.  In contrast, our method builds on an unconditional
SVBRDF diffusion backbone, trained independently from the incident lighting,
which can serve as a basis for conditional finetuning.  Furthermore, all the
above methods produce a single result per photograph, and offer no strategies
for producing alternative estimates that can better explain the appearance.

\paragraph{Iterative Inference Methods}
In contrast to direct inference methods that directly produce the target
material property maps, iterative inference methods perform an online
optimization to minimize a rendering loss with respect to the captured
photograph.  Gao~\etal~\shortcite{Gao:2019:DIR} and
Guo~\etal~\shortcite{Guo:2020:MRC} perform the optimization in a learned space
modeled by an auto-encoder and a GAN respectively. In both cases, the lighting
condition is only considered during the online optimization process, and the
space of SVBRDFs is lighting agnostic.  Hence, these methods could in theory
be applied to different lighting conditions. However, neither method provides
an interface for directing the optimization process to different plausible
SVBRDFs.  Furthermore, both methods tend to suffer from over-fitting,
resulting in burned-in highlights in the diffuse albedo maps.  Zhou and
Kalantari~\shortcite{Zhou:2022:LAT} and Fischer and
Ritschel~\shortcite{Fischer:2022:MML} combat overfitting by combing direct
inference and optimization-based methods using meta-learning.  While this
greatly improves the quality, the resulting trained networks are lighting
specific.  Our method is also iterative, but unlike the above methods, we do
not minimize a render loss function, but instead solve a denoising
differential equation. Unlike prior iterative methods, our method can produce
different replicate SVBRDFs by changing the input seed.

\paragraph{Generative Methods}
Aittala~\etal~\shortcite{Aittala:2016:RMN} extend parametric texture synthesis
to replicate the spatially varying appearance of a mostly stationary material
from a single flash lit photograph of an exemplar material. Similarly,
Wen~\etal~\shortcite{Wen:2022:SRS} train a GAN to model the appearance from a
photograph of a stationary material.
Henzler~\etal~\shortcite{Henzler:2021:GMB} employ a convolutional neural
network, conditioned on a latent code from a learned space, to convert a
random noise field into a random non-repeating field of BRDFs that match the
appearance of a flash-lit photograph of a stationary material.  Inspired by
MaterialGAN~\cite{Guo:2020:MRC}, Zhou~\etal~\shortcite{Zhou:2022:TGT} and
Hu~\etal~\shortcite{Hu:2022:CMA} introduce tileable material GANs that allow
for spatial control through an additional guidance image.  While these
networks can produce some stochastic variations around the expected value,
they do not effectively sample the distribution conditioned on the input
image. In contrast our method samples the conditional SVBRDF distribution that
better adheres to the input material's appearance.  An alternative strategy to
directly synthesizing the SVBRDF property maps, is to generate a procedural
model~\cite{Shi:2020:MDM,Hu:2022:AIP,Guerrero:2022:MAG}. The parameters of
such procedural models can be matched to the appearance of an exemplar in a
photograph~\cite{Guo:2020:ABI}. However, current procedural methods are
limited to specific material classes.

\section{SVBRDF Diffusion Model}
\label{sec:model}

\paragraph{Preliminaries} We model the appearance of a planar
spatially varying material by an SVBRDF, where each surface point's reflectance
is modeled by a microfacet BRDF with a GGX distribution~\cite{Walter:2007:MMR}
parameterized by its diffuse albedo, specular albedo, and monochrome specular
roughness.  In addition, we model the local surface variations by a normal map.

\paragraph{MatFusion} We first model the distribution of SVBRDFs using an
unconditional diffusion model, named MatFusion, that we will subsequently
refine based on the capture conditions.  The basic observation of diffusion
modeling is that adding noise to a signal (\eg, image) is a destructive
process, and hence the process of removing noise must therefore be generative.
In the limit, an entirely synthetic signal can be generated by starting from
pure random Gaussian noise, and iteratively denoising the
signal~\cite{Ho:2020:DDP}.  Formally, the goal of a generative model is to
sample a random variable according to a target data distribution $\x_0 \sim
\pdata$. In a diffusion model, we consider a sequence of related random
variables $\x_{1,2,\dots,\T}$ where each subsequent variable is increasingly
more noisy until $\x_\T$ is indistinguishable from pure Gaussian noise:
\begin{equation}
  \p(\x_\t | \x_0) = \N \left(\x_0, \s_\t^2\right),
\end{equation}
with $\sigma_\t > \sigma_{\t-1}$.  The diffusion process itself repeatedly
samples $\p(\x_{\t-1}|\x_\t)$ starting with $\t=\T$ and ending when
$\t=0$~\cite{Ho:2020:DDP,Song:2021:DDI}. This differs from a traditional
generator (\eg, GAN) that samples $\x_0$ directly.
Song~\etal~\shortcite{Song:2021:SBG} formulate diffusion as a differential
equation that maintains the distribution $\p$ as $\x$ evolves over time. The
change in $\x$ with time $\t$ is then\footnote{We assume no time-dependent
signal scaling, \ie, s(\t) = 1.}:
\begin{equation}
  \d\x = -\dot{\s}(\t) \s(\t) \nabla_\x \log \p(\x; \s(\t)) \d\t,
\label{eq:grad}
\end{equation}
where $\dot{\s}(\t)$ denotes the time-derivative of $\s(\t)$.
$\nabla_\x \log\p(\x; \sigma(\t))$ is also called the score function: a vector
that points towards the highest density of probable signals.  The differential
denoising equation can then be solved by taking discrete time-steps to evolve
the solution (\eg, using an Euler method) using \autoref{eq:grad}.  To compute
the score function, we define a neural denoising network $\D(\x_t; \t)$ that
minimizes the expected error on samples drawn from $\pdata$ for every
$\s_\t$. To avoid that the inputs of $\D$ grow with increasing $\s_\t$, it is
standard practice to normalize the estimate $\x_\t$ by
$\sqrt{1+\s^2_\t}$. Denoting the normalization factor of $\x_t$ as $\aw$,
abstracts the network input $\y$ as $\aw x + \bw n$ s.t.  $\aw^2 + \bw^2 = 1$,
where $\n$ is Gaussian distributed noise \footnote{$\aw$ and $\bw$ in this
  case correspond to $\sqrt{\bar{\alpha}}$ and $\sqrt{1 - \bar{\alpha}}$
  in~\cite{Ho:2020:DDP}.}.
Karras~\etal~\shortcite{Karras:2022:EDS} introduced a robust diffusion
variant, named k-diffusion, that instead of estimating the noise as in prior
diffusion models, estimates the ``velocity'' $\aw \n - \bw \x$ (note the
swapped position of $\n$ and $\x$ and change of sign for the second term) such
that the denoising network $\D$ minimizes the loss function:
\begin{equation}
 \E_{\x \sim \pdata} \, \E_{\n \sim \N(0, 1)} \, \left\lVert \D \left(\y; \t\right) - (\aw \n - \bw \x) \right\rVert_2^2,
\end{equation}
This allows us to estimate both the expectation of noise and signal with equal
ease by leveraging that $\aw^2 + \bw^2 = 1$:
\begin{eqnarray}
  \E_\n \approx \bw \y + \aw \D(\y; \t), \\
  \E_\x \approx \aw \y - \bw \D(\y; \t). \label{eq:kdiffx}
\end{eqnarray}
Note that depending on $\aw$ (which depends on $\s(\t)$), the output of the
neural network $\D$ varies from an estimate of the signal $\x$ to and estimate
of the noise $\n$ when $\t \rightarrow 0$.

\begin{figure}[!t]
  \centering
  \includegraphics[width=\linewidth]{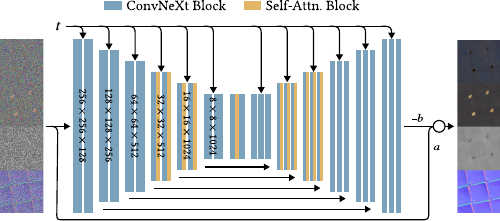}
  \caption{Summary of the MatFusion architecture.}
  \label{fig:architecture}
\end{figure}

\paragraph{Architecture}
In this paper we follow the normalization and sampling schedule (\ie,
$\s(\t)$) from~\cite{Ho:2020:DDP}, but use the k-diffusion loss function for
$\D$.  Our architecture for $\D$ is inspired by
Dhariwal~\etal~\shortcite{Dhariwal:2021:DMB}'s ImageNet-$256$ U-net
architecture with $6$ resolutions for the encoder and decoder
(\autoref{fig:architecture}).  To accommodate for the larger number of
channels ($10$ for SVBRDFs vs. $3$ for images), we employ a $3 \times 3 \times
10 \times 128$ convolution kernel to transform the $10$ input channels into
$128$ features.  We replace the Residual convolution blocks with ConvNeXt
blocks~\cite{Liu:2022:ACN} to increase the number of activations for the same
number of parameters; we argue that the higher channel count benefits from
more activations.  We follow DDIM~\cite{Song:2021:DDI} and encode $\t$ as a
$512$-length feature (using Fourier embedding and a 2-layer MLP) and pass it
to each ConvNeXt block as a dense residual layer between the $7 \times 7$
convolution and the first depth-wise convolution. Similar to DDIM, all layers
use a group norm with $32$ groups, and the $32$ and $16$ resolution layers
include self-attention blocks (with $8$ heads) after each ConvNeXt block, as
well as an additional attention-layer at the bottleneck.  We follow the method
of Rabe~\etal~\shortcite{Rabe:2021:SAD} to reduce the memory overhead of the
attention layers during training.

\paragraph{Conditional SVBRDF Diffusion Model}
In order to recover a plausible SVBRDF from a photograph, we need to make the
SVBRDF diffusion backbone network conditional on the photograph.  One possible
strategy to condition the neural network $\D$ on additional input images is by
concatenating them to the input
noise~\cite{VonPlaten:2022:DSA,Saharia:2022:PII}. However, this would require
retraining the diffusion network from scratch which is very
costly. Vonyov~\etal~\shortcite{Voynov:2022:SGT} perform sketch-guided
text-to-image diffusion by backpropagating the loss over the condition and an
inverse mapping from the diffusion output to the condition. In the context of
SVBRDFs, this would be akin to driving the diffusion process by the render
error, risking burn-in artifacts.  Recently, Zhang and
Agrawala~\shortcite{Zhang:2023:ACC} showed that an existing unconditional
diffusion model can be conditioned by adding zero-initialized dense layers to
each skip connection, and providing them the outputs of a parallel control
network trained on the conditional task.

Inspired by Zhang and Agrawala~\shortcite{Zhang:2023:ACC}, we expand the
\emph{input head} with $k$ additional features with both weights and bias
initialized with zeros (\ie, yielding an initial convolution kernel of
$3 \times 3 \times (10 + k) \times 128$, and where $k = 3 N$, and $N$ is the
number of condition input photographs).  Next, we \emph{finetune} the backbone
model for the target type of input photographs (unlike direct concatenation
which requires retraining from scratch).  Compared to ControlNet, our approach
is easier to implement and incurs less overhead as we do not need an
additional control network (we only expand the input head) at the cost of
``polluting'' the original diffusion network.

\begin{figure}[!t]
  \centering
  {\small
  \def\reswidthA{0.3\linewidth}
  \begin{tabular}{ccc}
    Input & First Step $\E_\x$ & Full Diffusion \\
    \includegraphics[width=\reswidthA]{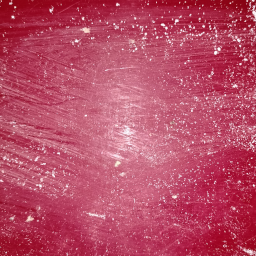} &
    \includegraphics[width=\reswidthA]{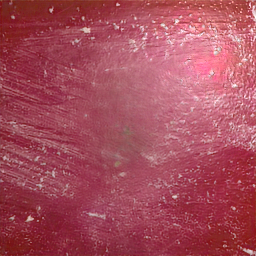} &
    \includegraphics[width=\reswidthA]{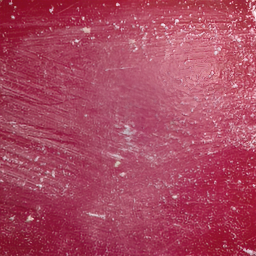} \\

    &
    \includegraphics[width=\reswidthA]{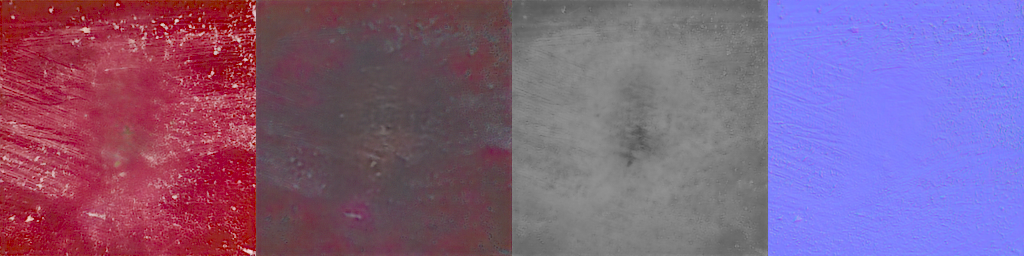} &
    \includegraphics[width=\reswidthA]{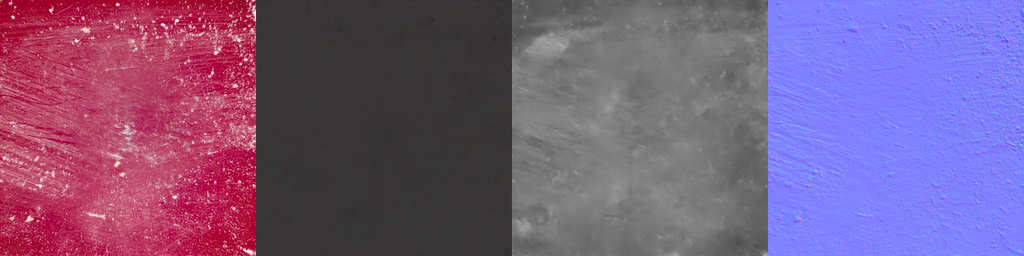}
  \end{tabular}
  }
  \caption{For the first diffusion step, the denoising neural network $\D$
    fully relies on the input photograph (left) and acts as a direct inference
    network (middle).  However, in contrast to direct inference, a diffusion
    model iteratively improves the estimate (right) by reducing burn-in,
    adding detail in the normal map, and improving diffuse-specular
    separation.}
  \label{fig:singlestep}
\end{figure}

\paragraph{Relation to Direct Inference}
When the k-diffusion model is conditioned on a photograph $\c$ of the target
material, the model subsumes direct inference methods. At $\t = \T$, the
signal $\y = \aw \x + \bw \n$ is purely Gaussian noise (\ie, $\aw \sim 0$),
and hence $\D(\y| \c; \t)$ mostly relies on the condition $\c$ to estimate the
velocity (\ie, $\aw \n - \bw \x \sim \x$).  For all practical purposes, we can
ignore the noisy input at $\t = \T$, and thus the \emph{expectation} $\E_\x$
computed from the estimate of $\D$ (\autoref{eq:kdiffx}) closely mimics the
behavior of a direct inference method. However, unlike direct inference
methods, diffusion only takes a small step towards the estimate and continues
to improve the result in subsequent steps. \autoref{fig:singlestep}
demonstrates that the expectation from the first diffusion step is similar to
the result of a direct inference method; note all SVBRDF property maps shown
in this paper are ordered as: diffuse albedo, specular albedo, roughness,
normal map. This initial estimate often exhibits burn-in, bended normals and
missing details, and imprecise diffuse-specular separation, which are reduced
in subsequent diffusion steps.

\section{Training Data}
\label{sec:training}

The MatFusion backbone model has $256$M parameters, hence, training such a
model requires a large and diverse training set.
Deschaintre~\etal~\shortcite{Deschaintre:2018:SIS} augment $150$ synthetic
SVBRDFs to $199,\!068$ training exemplars by randomly perturbing parameters,
scaling/rotating the exemplars, and taking convex combinations.  However,
since the dataset is augmented from only $150$ SVBRDFs, the texture diversity
is limited and insufficient to train our MatFusion backbone model.  To
mitigate this issue, we collected and augment $307$ additional synthetic
SVBRDFs from \url{https://polyhaven.com} and $1,\!570$ additional synthetic
SVBRDFs from \url{https://ambientcg.com}.

The $307$ SVBRDFs from Polyhaven are CC0 licensed and each contains a unique
diffuse albedo map, normal map and roughness map at $2k$
resolution. Polyhaven's SVBRDFs do not come with a specular albedo. We
therefore assign a homogeneous specular albedo uniformly sampled in
$[0.04, 0.08]$. The $1,\!570$ SVBRDFs from AmbientCG are also CC0 licensed,
and all contain unique albedo, specular roughness, and normal maps at $2k$
resolution. $274$ SVBRDFs also contain a metalness map.  A homogeneous
specular albedo is assigned (uniform random in $[0.04, 0.08]$) plus albedo
times metalness (if available).  The diffuse albedo is set to the albedo
(scaled by one minus metalness if available).

For each of the $1,\!877$ SVBRDF maps we randomly crop $16$ square areas, each
from from a random position, rotation, and size (between $512$ and $1,\!400$
pixels fully contained within the original maps). Each cropped map was
bilinearly resized to $512\times512$ resolution, yielding a total of
$30,\!032$ basis SVBRDFs.  To further diversify the roughness maps, we
randomly select $6,\!000$ basis SVBRDFs, and blend their roughness maps with
procedurally generated maps.  We employ a randomly initialized dense neural
network that transforms each pixels' (diffuse + specular) albedo and height
(obtained by integrating the normal map~\cite{Queau:2018:NIS}) to a procedural
roughness value; see the supplemental material for more details.  Note, the
randomly initialized network is not optimized and it serves as a random
non-linear transformation of albedo and height to roughness.

To better mimic that real-world materials are often formed by piece-wise
constant combinations of different basis materials (\eg, metal and rust), we
create $83,\!065$ additional piece-wise constant mixtures from both the
$199,\!068$ INRIA SVBRDFs and the $30,\!032$ basis SVBRDFs. For $66\%$ we mix
two randomly selected SVBRDFs without replacement (\ie, each SVBRDF is only
used in one mixture material), and three SVBRDFs for the remaining $34\%$.  We
use a randomly initialized dense neural network (detailed in the supplemental
material) that transforms each pixels' (diffuse + specular) albedo and height
into a one-hot selection weight (for each of the two/three source SVBRDFs).
Similar as for the roughness generator, the randomly initialized network is
not optimized and it serves as a random non-linear transformation and
thresholding step. To avoid unnatural hard edges, we perform the mixing on
$2\times$ bilinearly upsampled randomly selected $288\times288$ crops from the
INRIA or basis SVBRDFs, and after mixing, (average) downsample again to
$288\times288$ resolution.

Combining the INRIA training set ($199,\!068$ at $288\times 288$ resolution),
our basis SVBRDF set ($30,\!032$ at $512\times 512$ resolution), and the
mixture set ($83,\!065$ at $288\times 288$ resolution) yields our final
training set with $312,\!165$ training exemplars.  In addition, we created a
test set of $50$ materials that consists of a selection of $31$ diverse
materials from the Deep Inverse Rendering~\cite{Gao:2019:DIR} test set, $6$
materials from the look-ahead meta-learning~\cite{Zhou:2022:LAT} test set,
$11$ from Polyhaven, and $2$ from AmbientCG. None of the test materials are
included in the training set.

\section{Results}
\label{sec:results}

\begin{figure}[!t]
  \centering
  {\small
  \def\reswidthB{0.2\linewidth}
  \begin{tabular}{cccc}
    Flash w/o GI & Flash w/ GI & Natural w/o GI  & Natural w/ GI \\
    \includegraphics[width=\reswidthB]{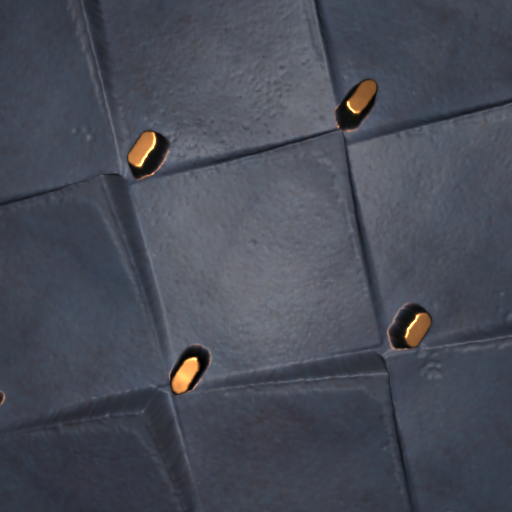} &
    \includegraphics[width=\reswidthB]{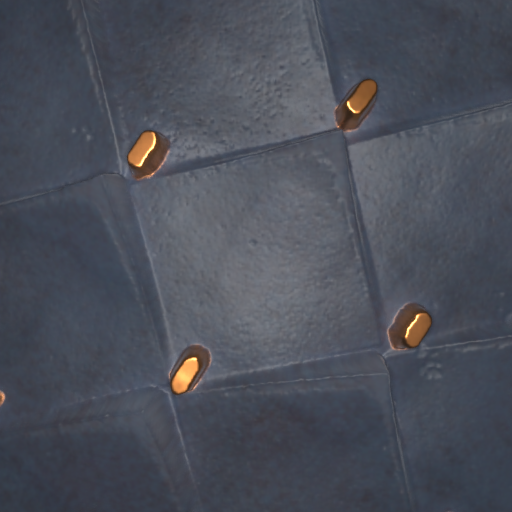} &
    \includegraphics[width=\reswidthB]{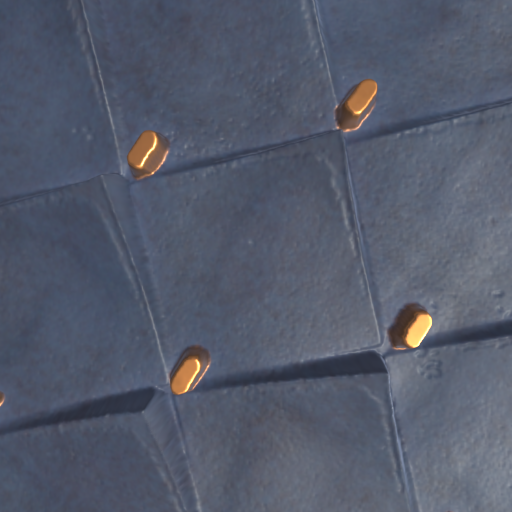} &
    \includegraphics[width=\reswidthB]{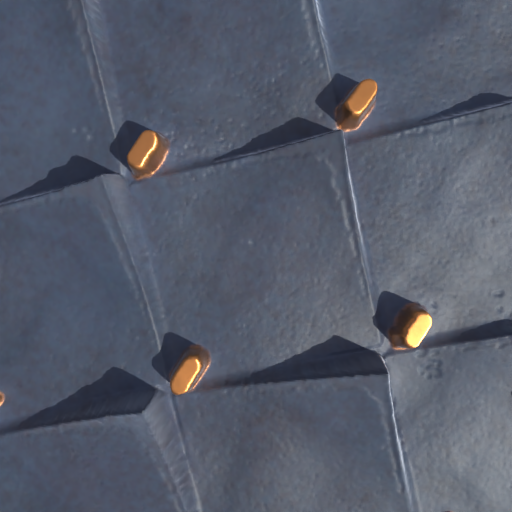}
  \end{tabular}
  }
  \caption{Global illumination transport within the spatially varying material
    is negligible for a colocated camera-light setup.  However, under natural
    lighting, the effects are significant (\ie, self-shadowing and ambient
    occlusion).}
  \label{fig:gi}
\end{figure}

\paragraph{Implementation} We implemented MatFusion in
FLAX~\cite{Heek:2023:FAN} and train it for $50$ epochs using the full
$312,\!165$ SVBRDF training set (cropped to $256 \times 256$ resolution) using
the AdamW optimizer~\cite{Loshchilov:2018:DWD} with a batch size of $32$, a
learning rate of $2 \times 10^{-5}$ (with a $100,\!000$ iteration warmup), and
EMA weights~\cite{Song:2020:ITT} on $4$ Nvidia A40 GPUs with $48$GB of memory.
Training took approximately $255$ hours.

We train three conditional variants of MatFusion. All three are finetuned for
$19$ epochs on MatFusion using the full SVBRDF training set using the same
optimizer and hyperparameters. Training took approximately $102$ hours on $4$
Nvidia A40 GPUs, or $2.5\times$ faster than training MatFusion from scratch.
The three variants differ in the expected lighting in the input condition
photograph: \Variant{colocated} flash lighting, \Variant{flash/no-flash}, and
\Variant{natural} lighting. The \Variant{Colocated} variant is trained on
synthetic photographs rendered with direct illumination only, as indirect
lighting is negligible for backscatter reflectance.  However, indirect
lighting significantly affects the appearance of spatially varying materials
(\autoref{fig:gi}).  Therefore, the \Variant{Natural} and
\Variant{Flash/no-flash} variants are trained on images rendered with
Blender's Cycles path-tracer with $32$ samples per pixel with OpenImageDenoise
using the height map as the material's geometry obtained by integrating the
surface normals~\cite{Queau:2018:NIS}; we use the original normal maps to
determine the shading normals.  Natural illumination is modeled by randomly
selecting and rotating an HDR environment map from $560$ CC0 licensed HDR
environment maps retrieved from \url{https://polyhaven.com/hdris}.  For the
\Variant{Flash/no-flash} variant, the \emph{log} relative brightness ratio
between the flash lighting and the environment lighting is randomly sampled
between $\log(1/50)$ and $\log(3/2)$.  Both the \Variant{Natural} and
\Variant{Flash/no-flash} variants are trained on images rendered with a
virtual camera with a focal length of $35$mm (\ie, camera distance = exemplar
size).  The \Variant{Colocated} variant is trained for a variable camera
distance (with matching FOV) sampled according to a $\frac{1}{2} \Gamma(2,2)$
distribution (relative to the exemplar size), and we concatenate the per-pixel
view vector as an additional input condition.

During inference, the differential equation is iteratively solved using the
EulerA solver~\cite{Song:2021:SBG} in just $20$ steps and with the guidance
scale set to $1$.

\begin{figure*}[!t]
  {\small
  \def\reswidthC{0.08\linewidth}
  \def\rsep{\hspace{0.1em}}
  \begin{tabular}{cc@{\rsep}c@{\rsep}c@{\rsep}c@{\rsep}c@{\rsep}c@{\rsep}c@{\rsep}c@{\rsep}c@{\rsep}c@{\rsep}}
    Input &
    \parbox[b]{\reswidthC}{\centering Fixed Seed} &
    &
    \parbox[b]{\reswidthC}{\centering Render Err.} &
    &
    &
    &
    &
    &
    &
    Manual \\
    \includegraphics[width=\reswidthC]{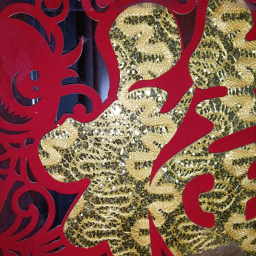} &
    \includegraphics[width=\reswidthC]{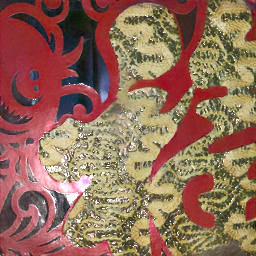} &
    \includegraphics[width=\reswidthC]{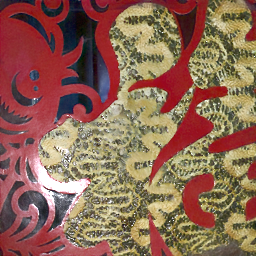} &
    \includegraphics[width=\reswidthC]{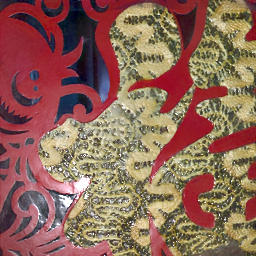} &
    \includegraphics[width=\reswidthC]{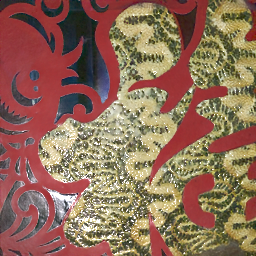} &
    \includegraphics[width=\reswidthC]{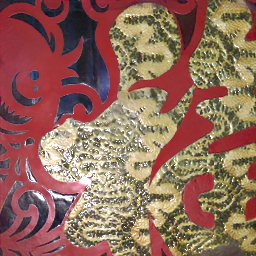} &
    \includegraphics[width=\reswidthC]{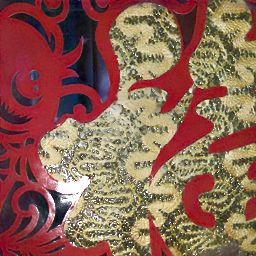} &
    \includegraphics[width=\reswidthC]{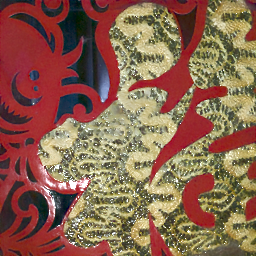} &
    \includegraphics[width=\reswidthC]{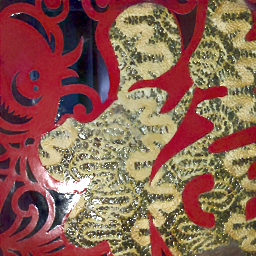} &
    \includegraphics[width=\reswidthC]{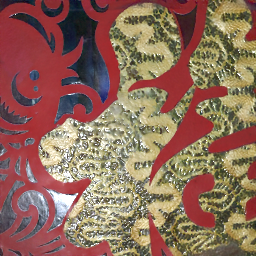} &
    \includegraphics[width=\reswidthC]{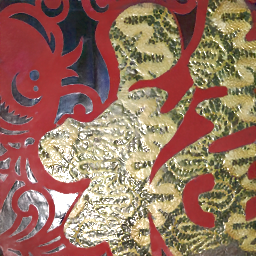} \\
    &
    \includegraphics[width=\reswidthC]{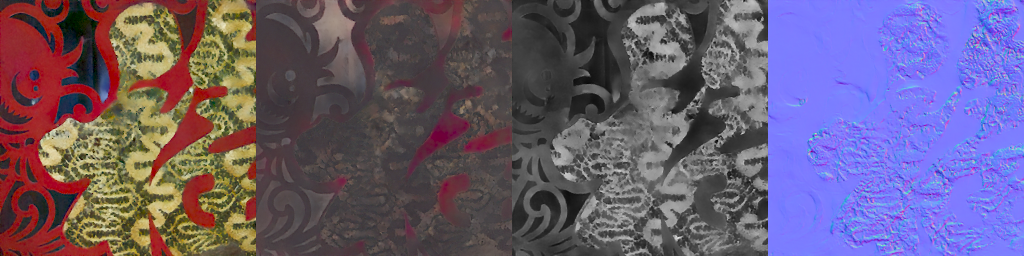} &
    \includegraphics[width=\reswidthC]{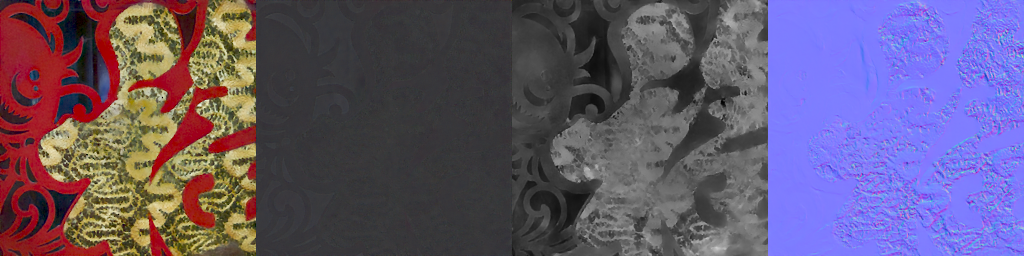} &
    \includegraphics[width=\reswidthC]{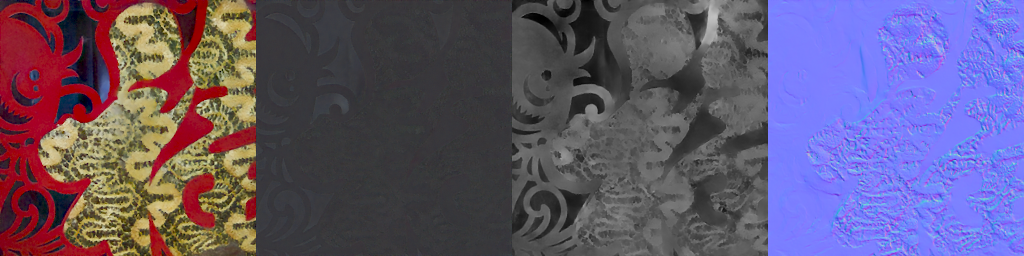} &
    \includegraphics[width=\reswidthC]{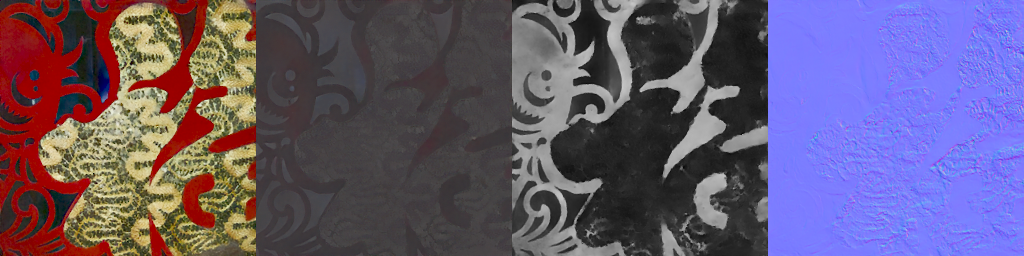} &
    \includegraphics[width=\reswidthC]{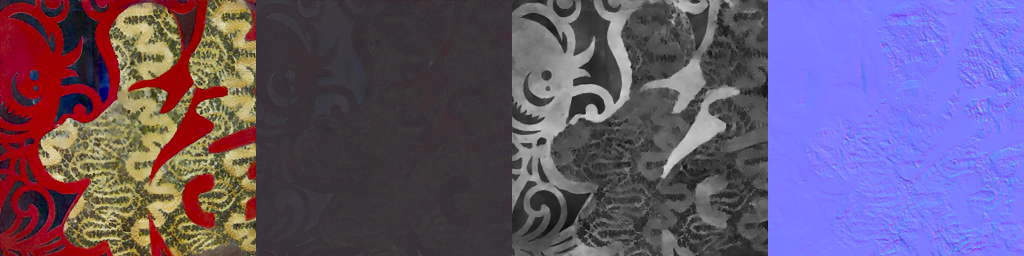} &
    \includegraphics[width=\reswidthC]{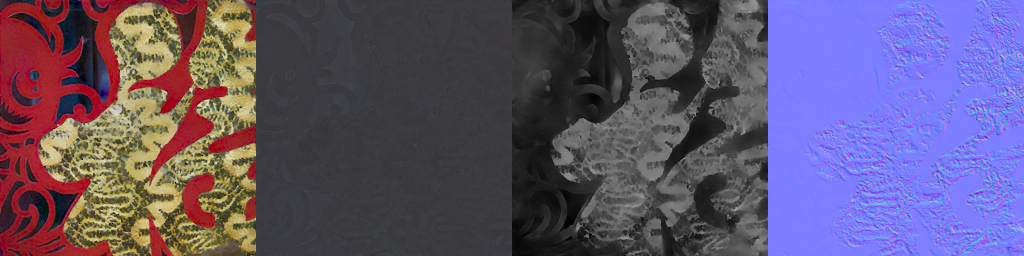} &
    \includegraphics[width=\reswidthC]{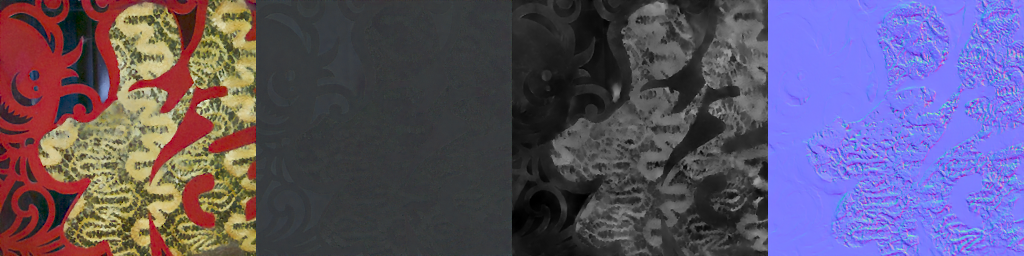} &
    \includegraphics[width=\reswidthC]{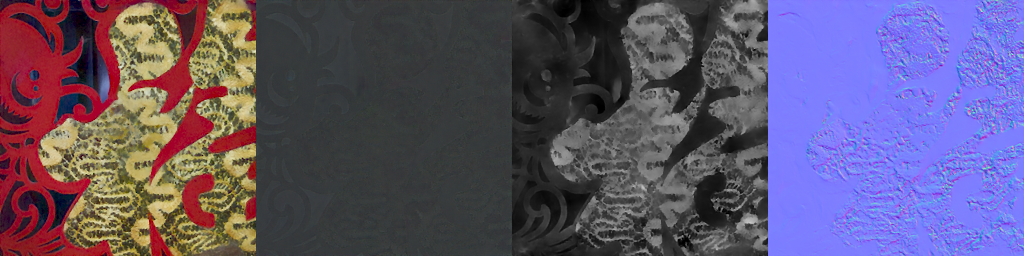} &
    \includegraphics[width=\reswidthC]{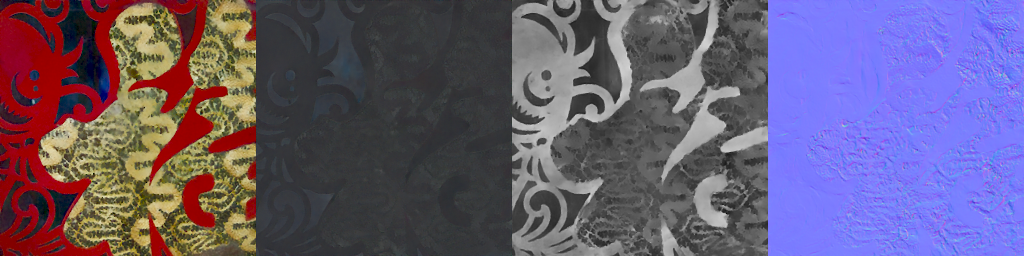} &
    \includegraphics[width=\reswidthC]{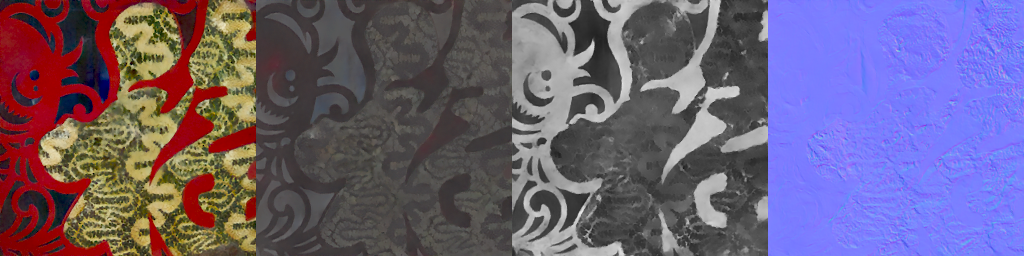} \\
  \end{tabular}
  }
  \centering
  \caption{Changing the seed results in different SVBRDF replicates
    conditioned on the input photograph. For each replicate we show a
    rendering under a different lighting than the input photograph as well as
    the generated SVBRDF property maps. Also marked are the SVBRDF selection
    based on the render error with respect to the input lighting, as well as
    the manual selection of the (subjectively) most plausible SVBRDF.}
  \label{fig:replicates}
\end{figure*}

\paragraph{Selection}
The conditional SVBRDF diffusion models take, besides the input photograph,
also a normal distributed random field determined by a seed.  By changing the
seed, different replicates of the SVBRDF can be generated
(\autoref{fig:replicates}).  The choice of the seed can impact the quality of
the result.  Therefore, we show results selected with one of the following
three selection strategies:
\begin{enumerate}
  \item \emph{Fixed seed}: the seed is fixed for all results.
  \item \emph{Render error selection}: we render the generated SVBRDFs from
    $10$ random seeds and select the one that minimizes the LPIPS
    error~\cite{Zhang:2018:TUE} when rendered under the capture lighting
    conditions.
  \item \emph{Manual selection}: a set of $10$ SVBRDFs generated with
    different random seeds are presented and the user manually selects the
    SVBRDF that appears (subjectively) the most plausible.
\end{enumerate}
We also experimented with optimizing the input random field on the render
error, but found that this tends to produce burn-in of the specular highlight.
While the majority of seeds do not produce burn-in, those that do are
scattered through the whole space. Thus no matter the starting point, there is
always a nearby point that produces burn-in which the optimization will
inevitably drive the solution towards.

\begin{figure*}
  {\small
  \def\reswidthG{0.117\linewidth}
  \def\resdwidthG{0.234\linewidth}  
  \def\rsep{\hspace{0.1em}}
  \begin{tabular}{c@{\rsep}cc@{\rsep}cc@{\rsep}cc@{\rsep}c}
    \multicolumn{2}{c}{Reference} &
    \multicolumn{2}{c}{\Variant{Colocated}} &
    \multicolumn{2}{c}{\Variant{Natural}} &
    \multicolumn{2}{c}{\Variant{Flash/no-flash}} \\

    \includegraphics[width=\reswidthG]{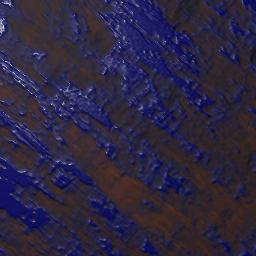} &
    \includegraphics[width=\reswidthG]{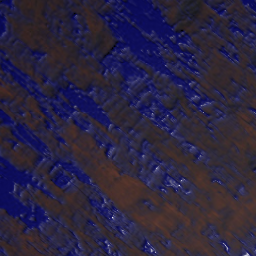} &
    \includegraphics[width=\reswidthG]{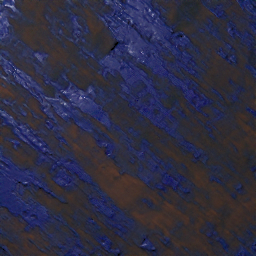} &
    \includegraphics[width=\reswidthG]{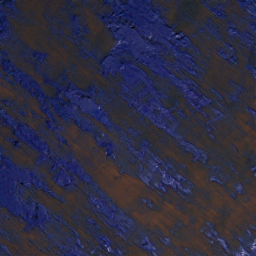} &
    \includegraphics[width=\reswidthG]{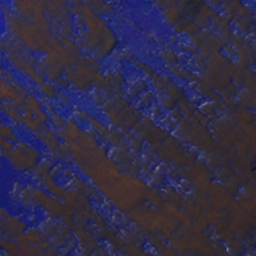} &
    \includegraphics[width=\reswidthG]{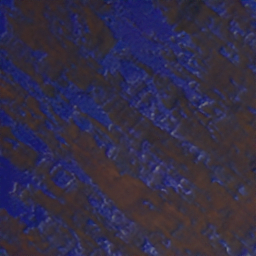} &
    \includegraphics[width=\reswidthG]{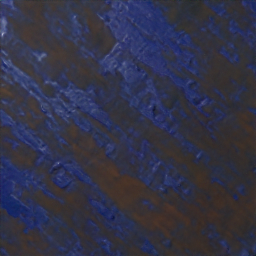} &
    \includegraphics[width=\reswidthG]{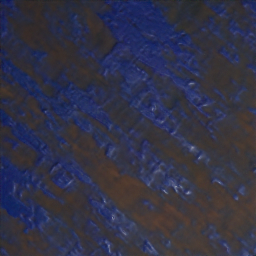} \\
    \multicolumn{2}{c}{\includegraphics[width=\resdwidthG]{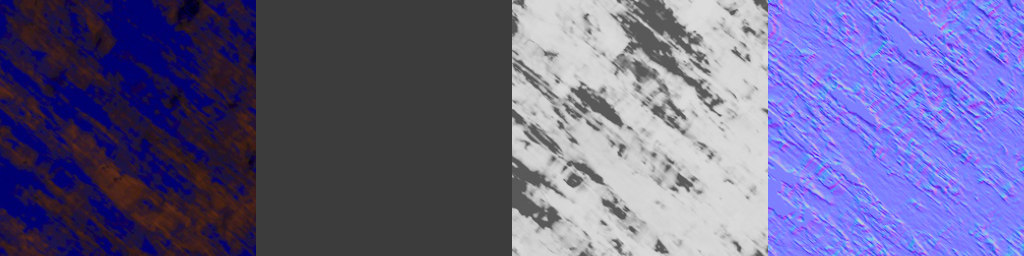}} &
    \multicolumn{2}{c}{\includegraphics[width=\resdwidthG]{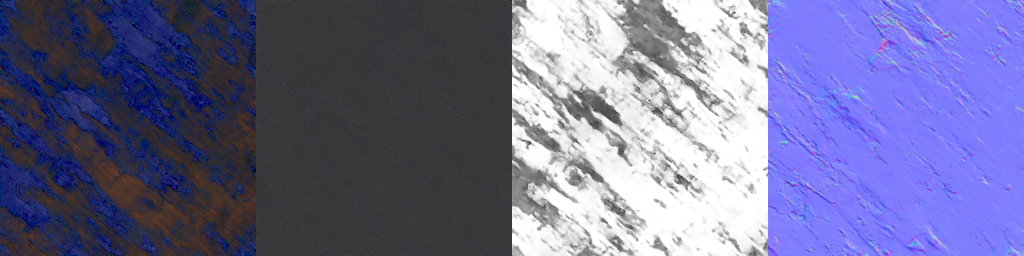}} &
    \multicolumn{2}{c}{\includegraphics[width=\resdwidthG]{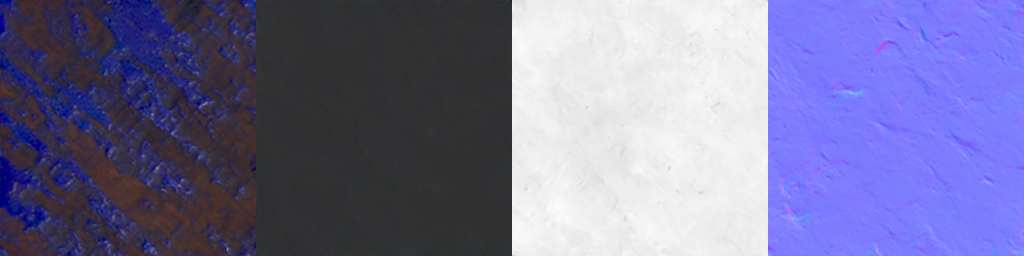}} &
    \multicolumn{2}{c}{\includegraphics[width=\resdwidthG]{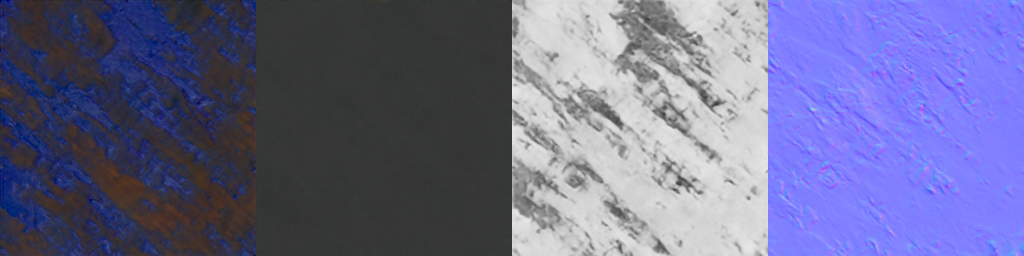}} \\

    \includegraphics[width=\reswidthG]{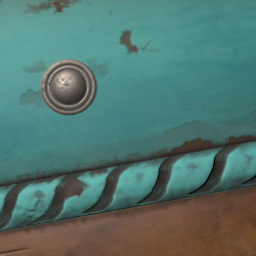} &
    \includegraphics[width=\reswidthG]{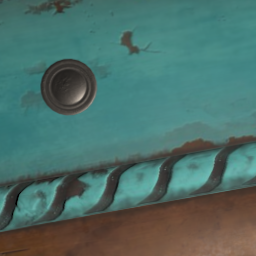} &
    \includegraphics[width=\reswidthG]{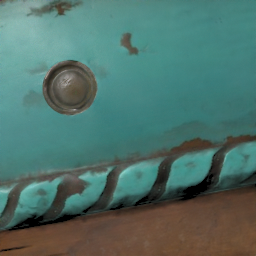} &
    \includegraphics[width=\reswidthG]{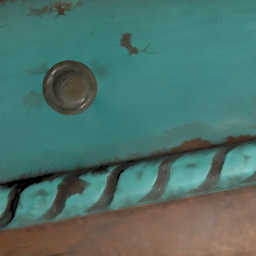} &
    \includegraphics[width=\reswidthG]{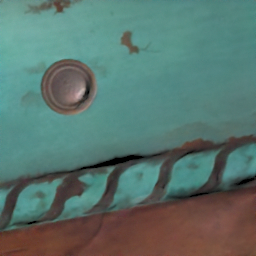} &
    \includegraphics[width=\reswidthG]{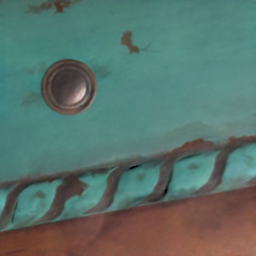} &
    \includegraphics[width=\reswidthG]{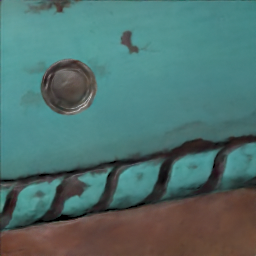} &
    \includegraphics[width=\reswidthG]{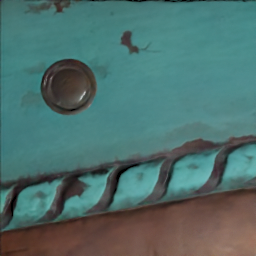} \\
    \multicolumn{2}{c}{\includegraphics[width=\resdwidthG]{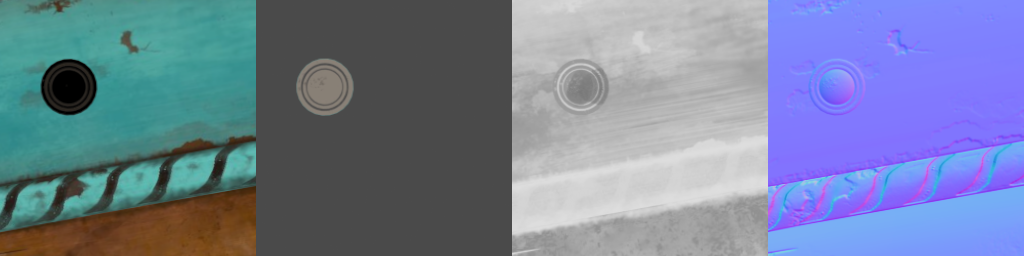}} &
    \multicolumn{2}{c}{\includegraphics[width=\resdwidthG]{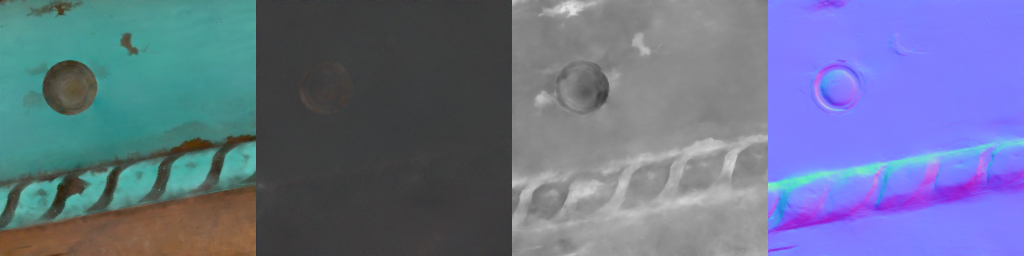}} &
    \multicolumn{2}{c}{\includegraphics[width=\resdwidthG]{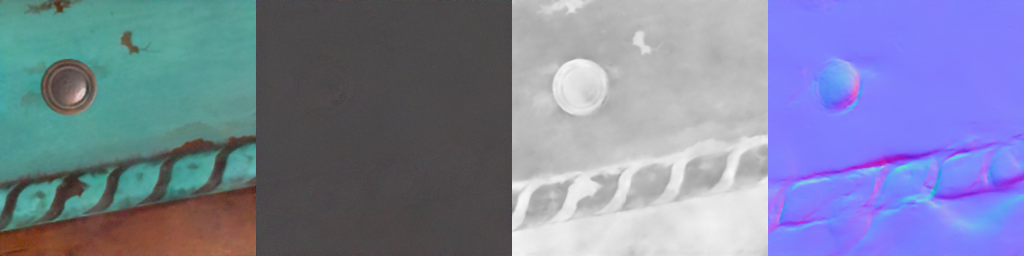}} &
    \multicolumn{2}{c}{\includegraphics[width=\resdwidthG]{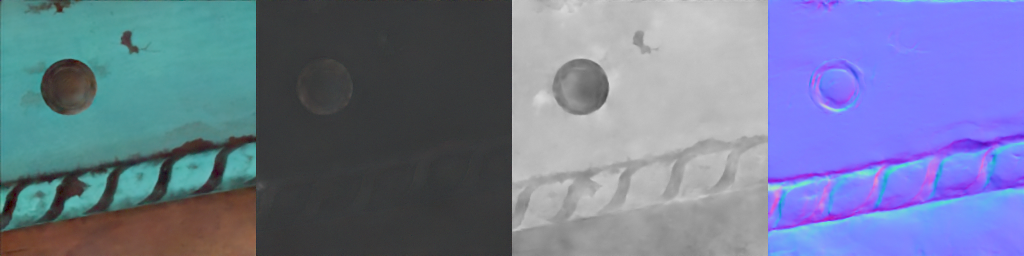}} \\

    \includegraphics[width=\reswidthG]{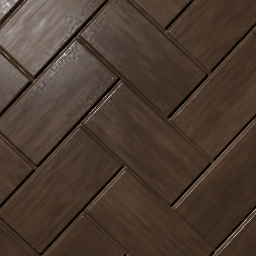} &
    \includegraphics[width=\reswidthG]{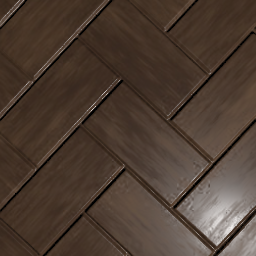} &
    \includegraphics[width=\reswidthG]{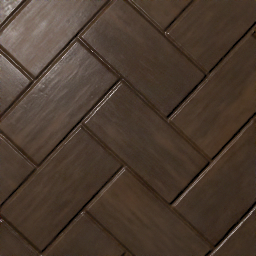} &
    \includegraphics[width=\reswidthG]{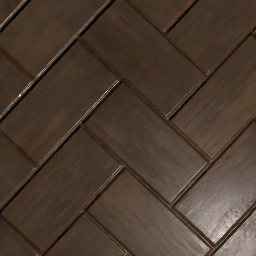} &
    \includegraphics[width=\reswidthG]{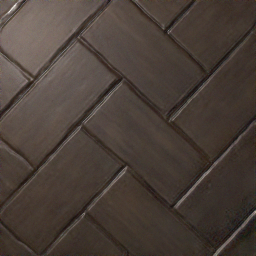} &
    \includegraphics[width=\reswidthG]{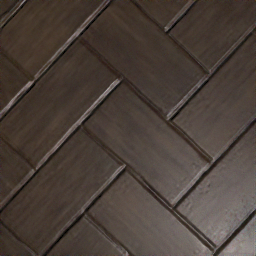} &
    \includegraphics[width=\reswidthG]{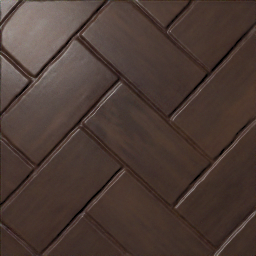} &
    \includegraphics[width=\reswidthG]{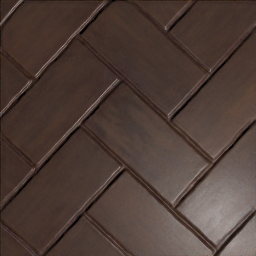} \\
    \multicolumn{2}{c}{\includegraphics[width=\resdwidthG]{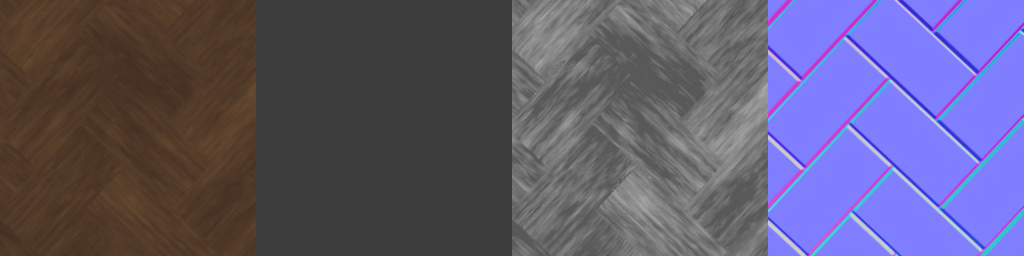}} &
    \multicolumn{2}{c}{\includegraphics[width=\resdwidthG]{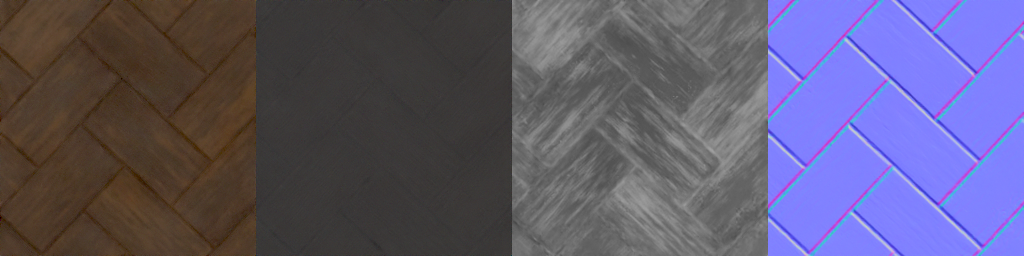}} &
    \multicolumn{2}{c}{\includegraphics[width=\resdwidthG]{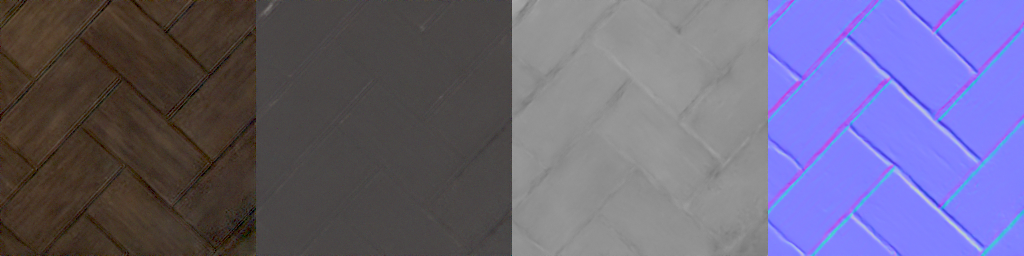}} &
    \multicolumn{2}{c}{\includegraphics[width=\resdwidthG]{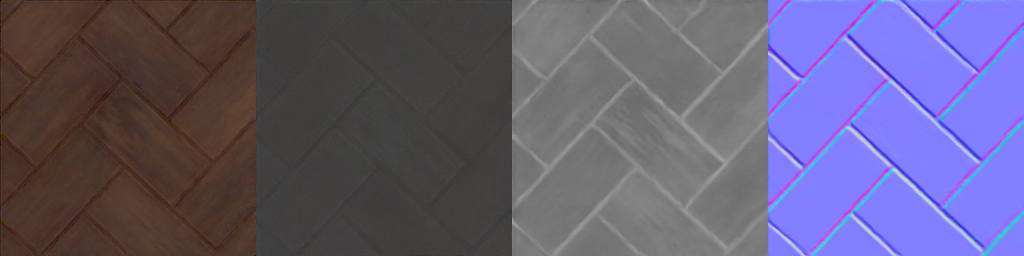}} \\

    \includegraphics[width=\reswidthG]{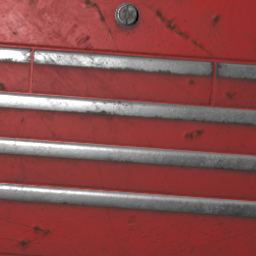} &
    \includegraphics[width=\reswidthG]{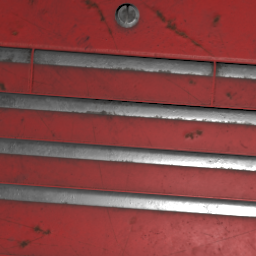} &
    \includegraphics[width=\reswidthG]{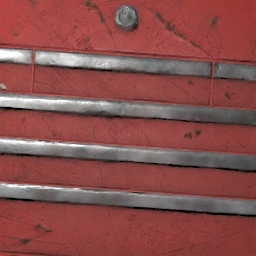} &
    \includegraphics[width=\reswidthG]{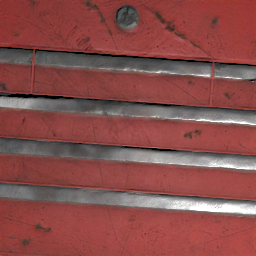} &
    \includegraphics[width=\reswidthG]{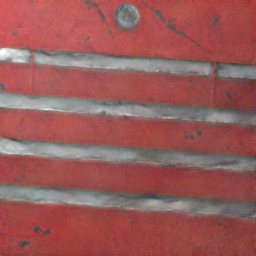} &
    \includegraphics[width=\reswidthG]{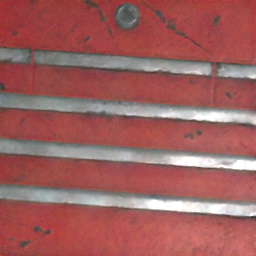} &
    \includegraphics[width=\reswidthG]{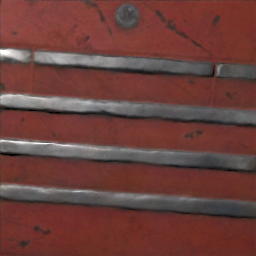} &
    \includegraphics[width=\reswidthG]{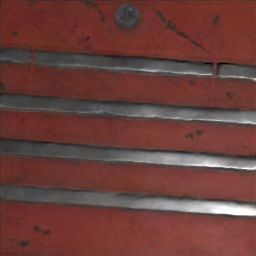} \\
    \multicolumn{2}{c}{\includegraphics[width=\resdwidthG]{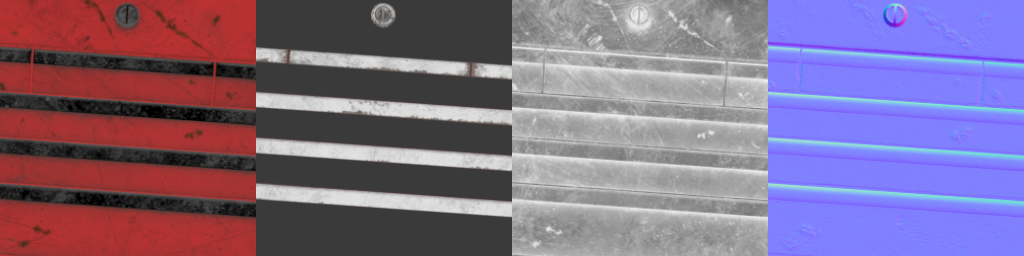}} &
    \multicolumn{2}{c}{\includegraphics[width=\resdwidthG]{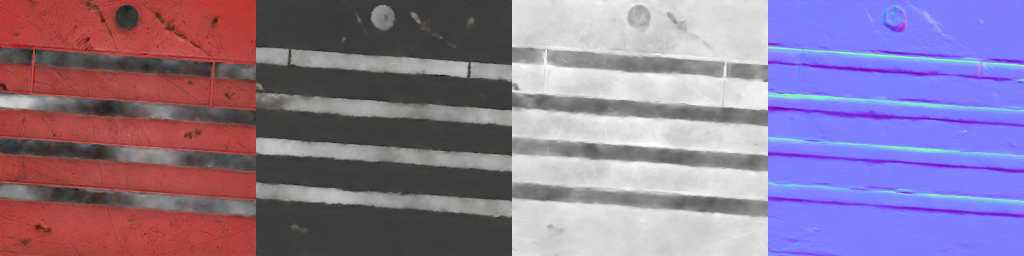}} &
    \multicolumn{2}{c}{\includegraphics[width=\resdwidthG]{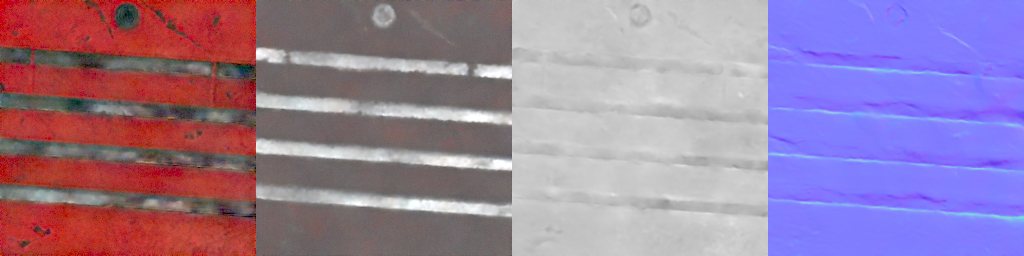}} &
    \multicolumn{2}{c}{\includegraphics[width=\resdwidthG]{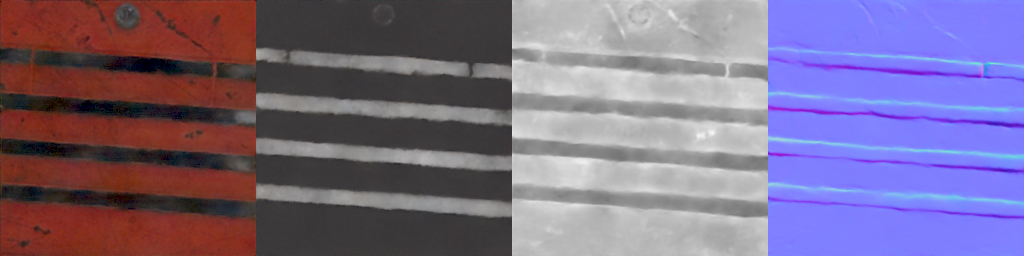}} \\

    \includegraphics[width=\reswidthG]{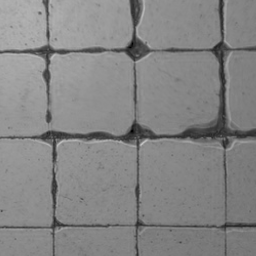} &
    \includegraphics[width=\reswidthG]{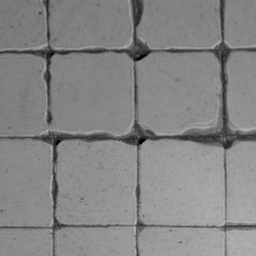} &
    \includegraphics[width=\reswidthG]{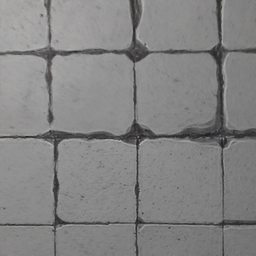} &
    \includegraphics[width=\reswidthG]{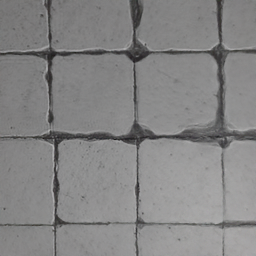} &
    \includegraphics[width=\reswidthG]{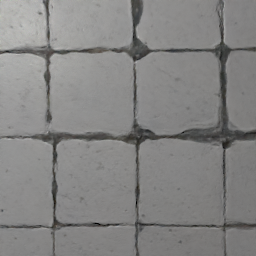} &
    \includegraphics[width=\reswidthG]{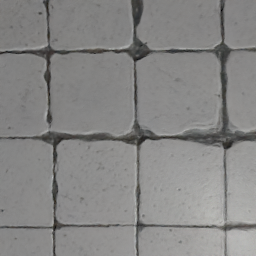} &
    \includegraphics[width=\reswidthG]{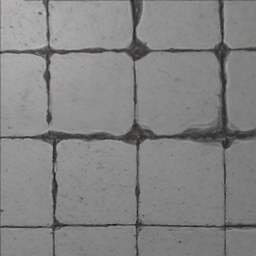} &
    \includegraphics[width=\reswidthG]{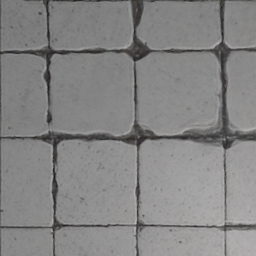} \\
    \multicolumn{2}{c}{\includegraphics[width=\resdwidthG]{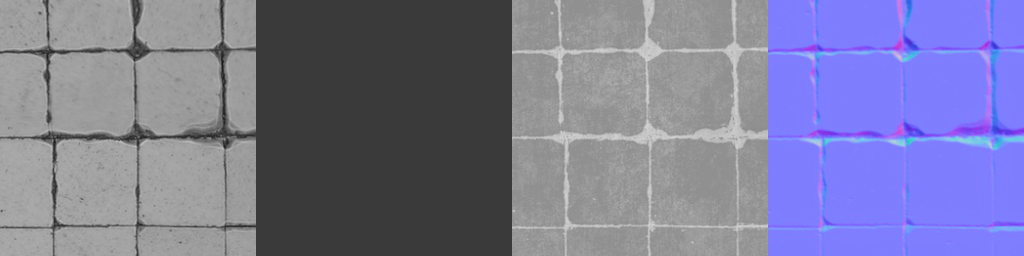}} &
    \multicolumn{2}{c}{\includegraphics[width=\resdwidthG]{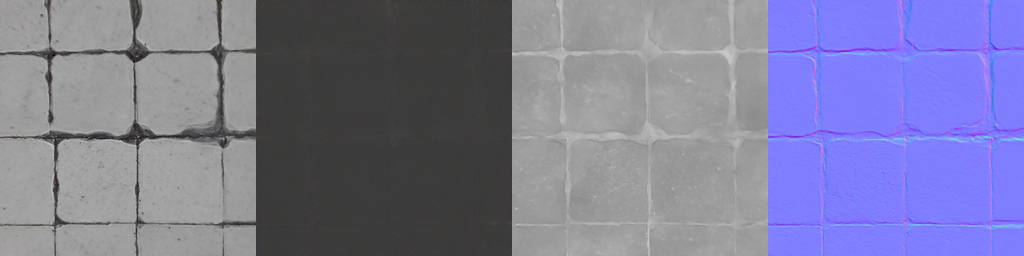}} &
    \multicolumn{2}{c}{\includegraphics[width=\resdwidthG]{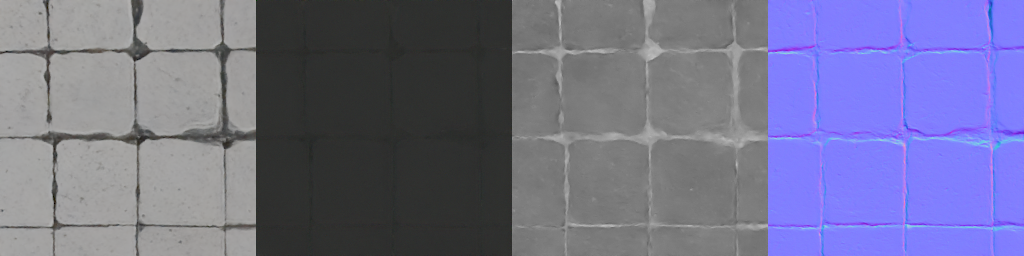}} &
    \multicolumn{2}{c}{\includegraphics[width=\resdwidthG]{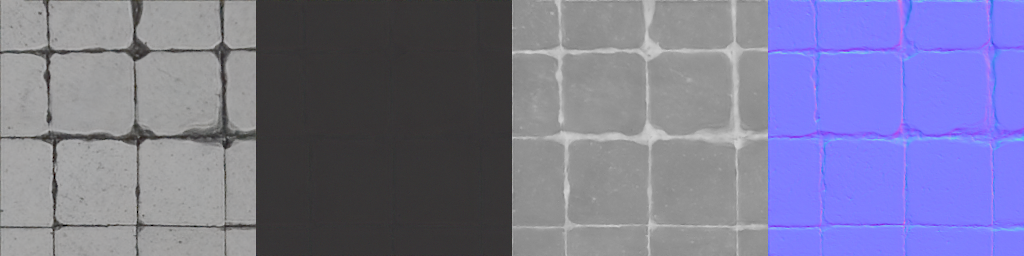}} \\

    \includegraphics[width=\reswidthG]{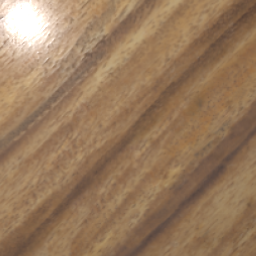} &
    \includegraphics[width=\reswidthG]{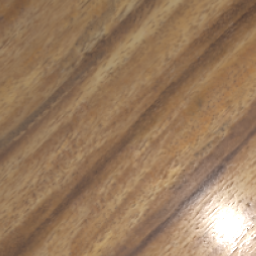} &
    \includegraphics[width=\reswidthG]{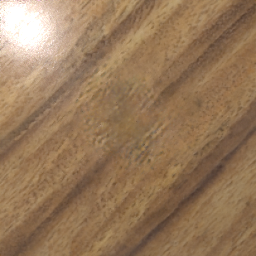} &
    \includegraphics[width=\reswidthG]{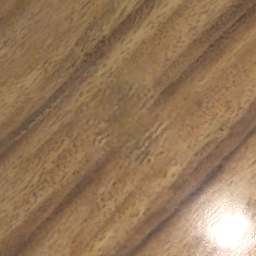} &
    \includegraphics[width=\reswidthG]{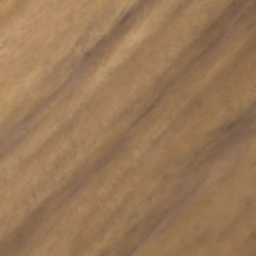} &
    \includegraphics[width=\reswidthG]{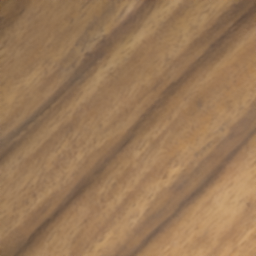} &
    \includegraphics[width=\reswidthG]{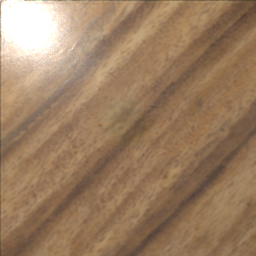} &
    \includegraphics[width=\reswidthG]{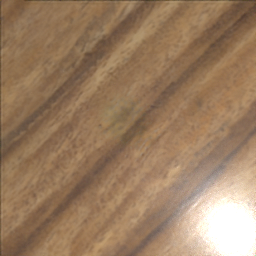} \\
    \multicolumn{2}{c}{\includegraphics[width=\resdwidthG]{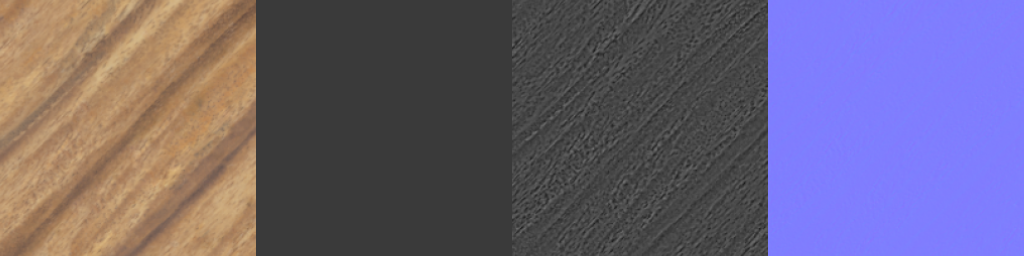}} &
    \multicolumn{2}{c}{\includegraphics[width=\resdwidthG]{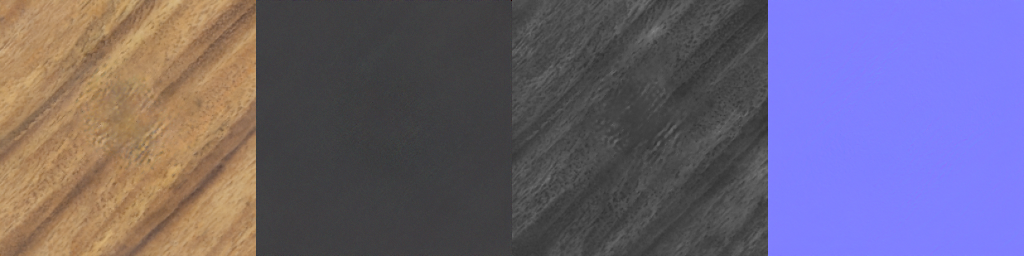}} &
    \multicolumn{2}{c}{\includegraphics[width=\resdwidthG]{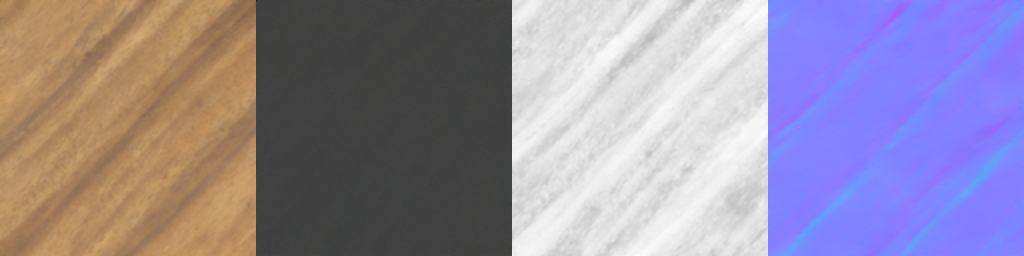}} &
    \multicolumn{2}{c}{\includegraphics[width=\resdwidthG]{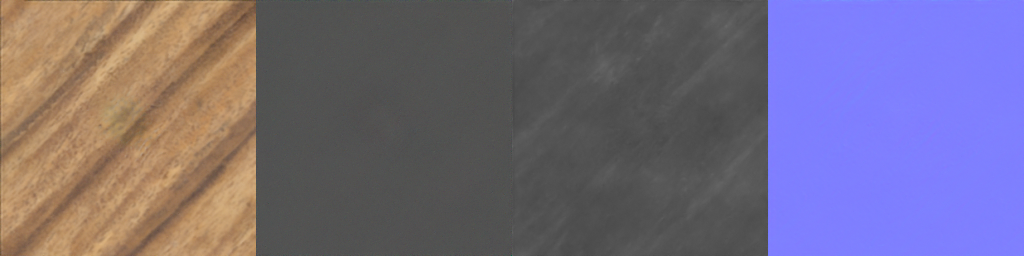}}
  \end{tabular}
  }
  \centering
  \caption{Comparison of the \Variant{Colocated}, \Variant{Natural}, and
    \Variant{Flash/no-flash} conditional diffusion models on a variety of
    synthetic SVBRDFs.}
  \label{fig:results}
\end{figure*}

\paragraph{Synthetic Results}
\autoref{fig:results} compares the estimated SVBRDFs, manually selected from
$10$ random seeds, for $6$ selected synthetic materials for each of the three
conditional diffusion models. For each material, we show two renderings under
different point lights for each of the models and the reference. In general,
the \Variant{colocated} model produces the most consistent results due to the
known lighting, although it sometimes fails to recover the specular
reflectance on small features (\eg, the nob in the 2nd material) or produces
unexpected texture variations (\eg, the center of the 6th material). The
results from the \Variant{natural} model exhibit a greater variability in
accuracy, such as incomplete diffuse-specular separation (4th example), or
underestimation of specular roughness (6th example).  Nevertheless, the
resulting SVBRDFs are still plausible, demonstrating the ability of MatFusion
to recover the SVBRDFs of general spatially varying materials under unknown
lighting.  The \Variant{Flash/no-flash} model benefits from having an input
without strong specular highlights (\ie, no-flash) to better recover the
diffuse texture. On the other hand, due to the unknown relative brightness of
the natural lighting versus the flash lighting, it sometimes underestimates
either the diffuse albedo (\eg, 4th material) or the specular roughness (\eg,
3rd material).  The \Variant{Flash/no-flash} model shows that MatFusion can be
conditioned on more than one input.

\begin{figure*}
  \centering
  {
  \def\reswidth{0.13\linewidth}
  \setlength{\tabcolsep}{0.6mm}
  \renewcommand{\arraystretch}{0.6}
  \begin{tabular}{ccccccc}
    Input &
    \multicolumn{1}{c}{Reference} &
    \multicolumn{1}{c}{Fixed Seed} &
    \multicolumn{1}{c}{Render Error} &
    \multicolumn{1}{c}{Manual} &
    \multicolumn{1}{c}{Zhou~\etal~\shortcite{Zhou:2021:ASI}} &
    \multicolumn{1}{c}{Zhou~\etal~\shortcite{Zhou:2022:LAT}} \\

    \includegraphics[width=\reswidth]{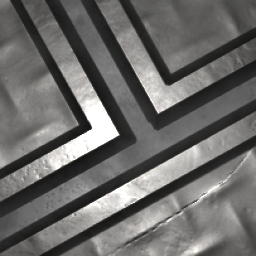} &
    \includegraphics[width=\reswidth]{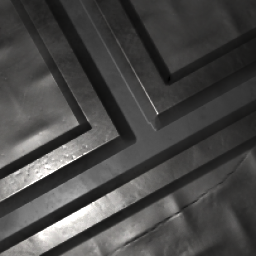} &
    \includegraphics[width=\reswidth]{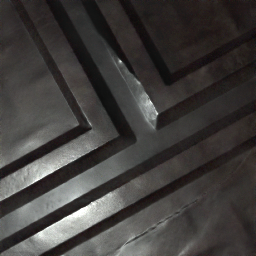} &
    \includegraphics[width=\reswidth]{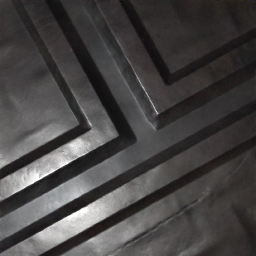} &
    \includegraphics[width=\reswidth]{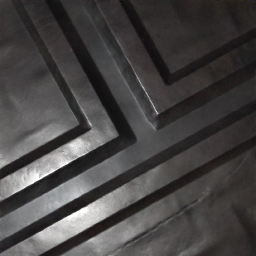} &
    \includegraphics[width=\reswidth]{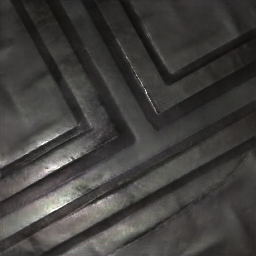} &
    \includegraphics[width=\reswidth]{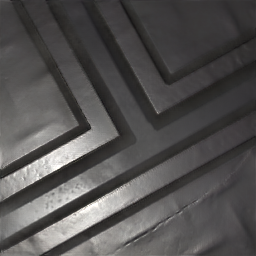} \\

    &
    \includegraphics[width=\reswidth]{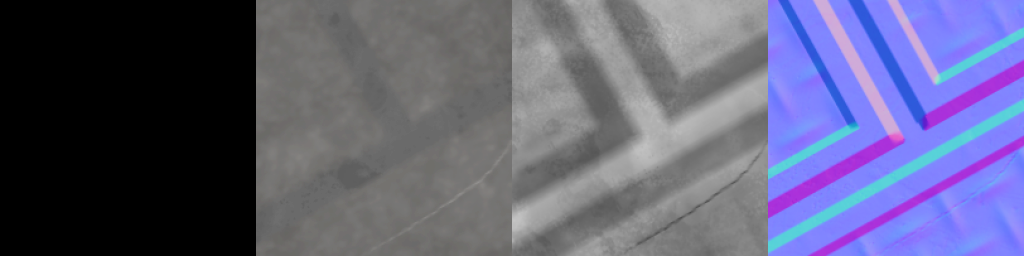} &
    \includegraphics[width=\reswidth]{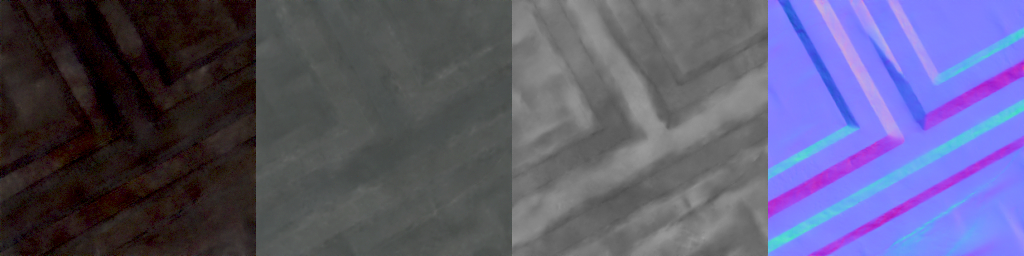} &
    \includegraphics[width=\reswidth]{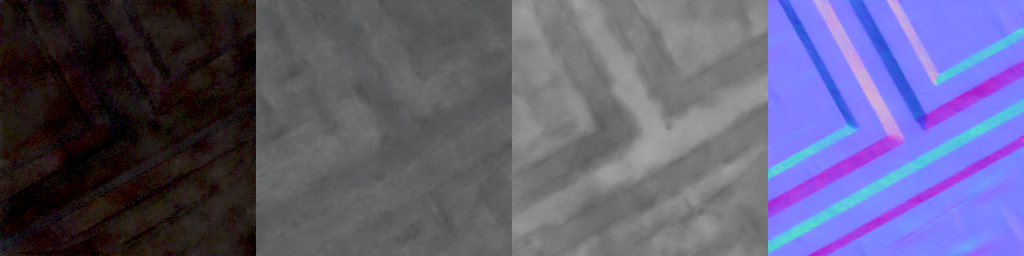} &
    \includegraphics[width=\reswidth]{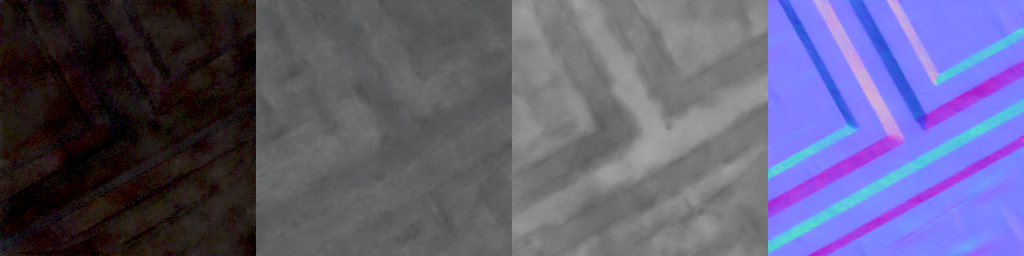} &
    \includegraphics[width=\reswidth]{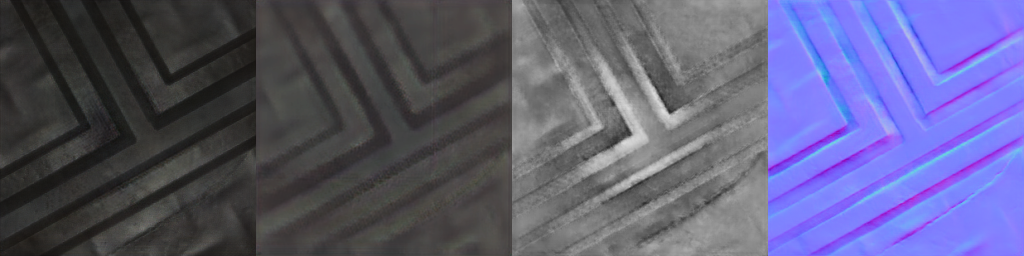} &
    \includegraphics[width=\reswidth]{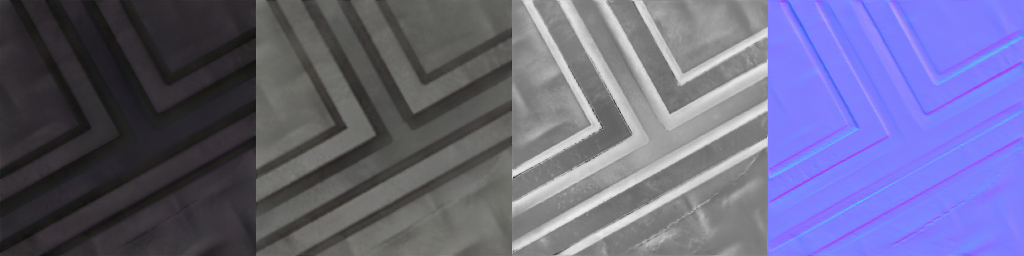} \\

    \multicolumn{2}{r}{Average LPIPS Render Error:} &
    0.1762 &
    0.0984 &
    0.0984 &
    0.2450 &
    0.2170 \\

    \includegraphics[width=\reswidth]{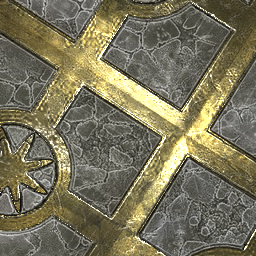} &
    \includegraphics[width=\reswidth]{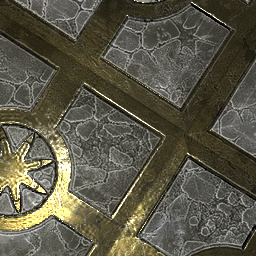} &
    \includegraphics[width=\reswidth]{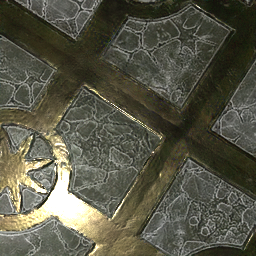} &
    \includegraphics[width=\reswidth]{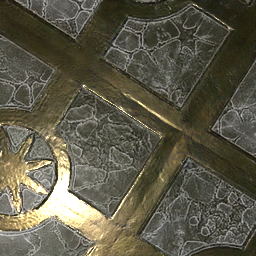} &
    \includegraphics[width=\reswidth]{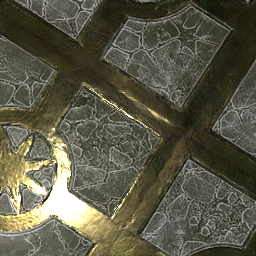} &
    \includegraphics[width=\reswidth]{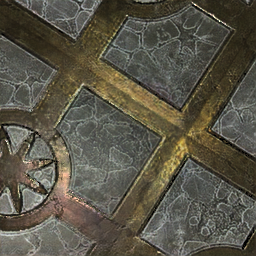} &
    \includegraphics[width=\reswidth]{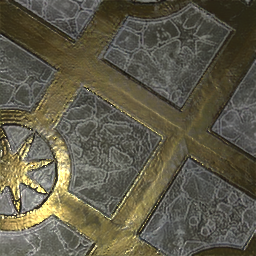} \\

    &
    \includegraphics[width=\reswidth]{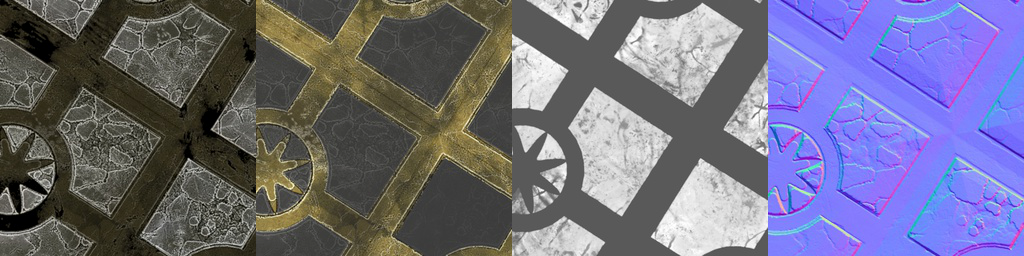} &
    \includegraphics[width=\reswidth]{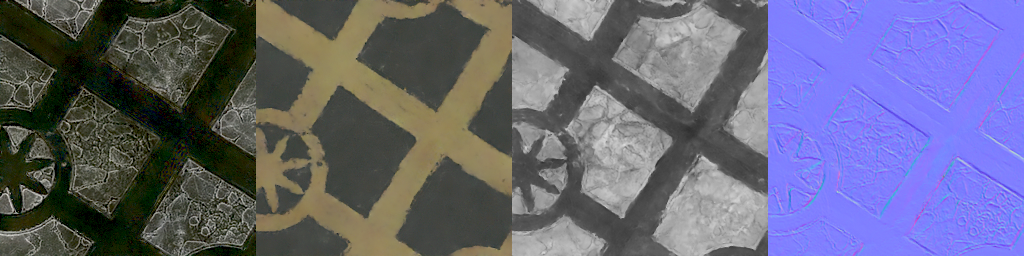} &
    \includegraphics[width=\reswidth]{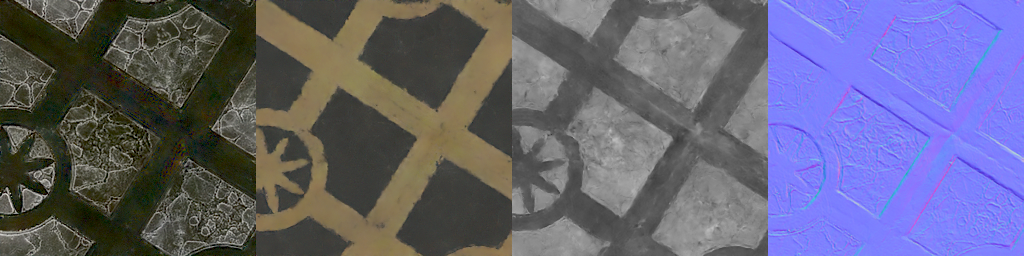} &
    \includegraphics[width=\reswidth]{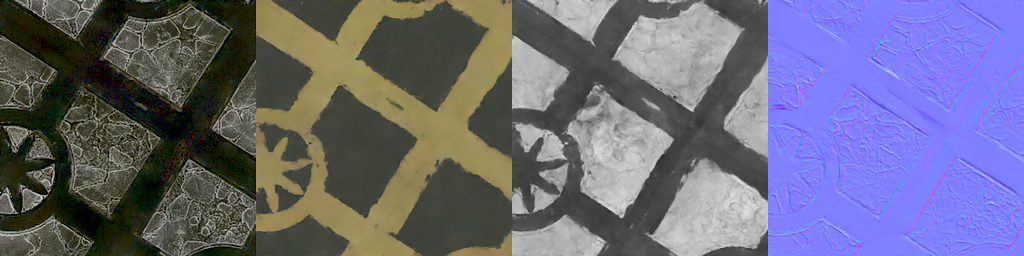} &
    \includegraphics[width=\reswidth]{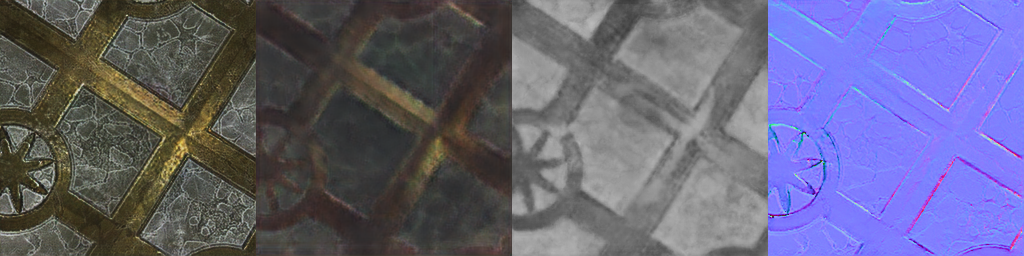} &
    \includegraphics[width=\reswidth]{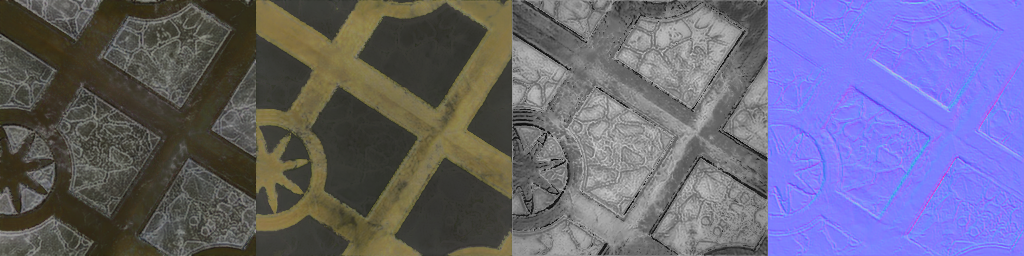} \\

    \multicolumn{2}{r}{Average LPIPS Render Error:} &
    0.2324 &
    0.2391 &
    0.2146 &
    0.2930 &
    0.3076 \\

    \includegraphics[width=\reswidth]{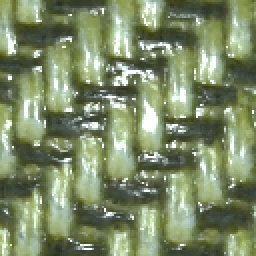} &
    \includegraphics[width=\reswidth]{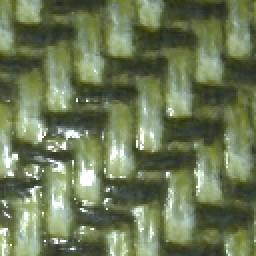} &
    \includegraphics[width=\reswidth]{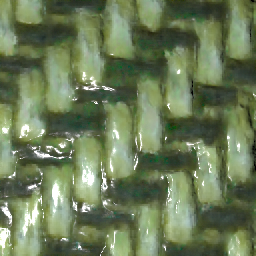} &
    \includegraphics[width=\reswidth]{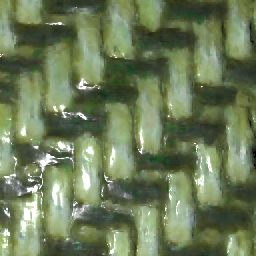} &
    \includegraphics[width=\reswidth]{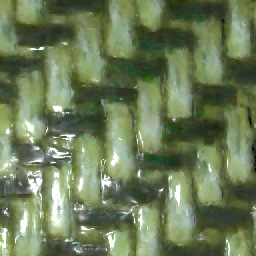} &
    \includegraphics[width=\reswidth]{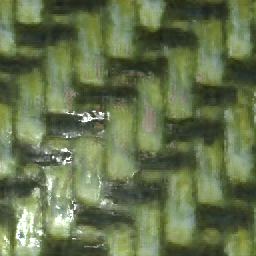} &
    \includegraphics[width=\reswidth]{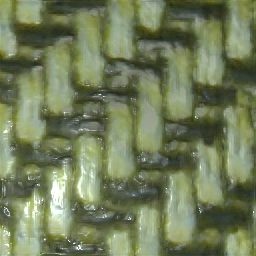} \\

    &
    \includegraphics[width=\reswidth]{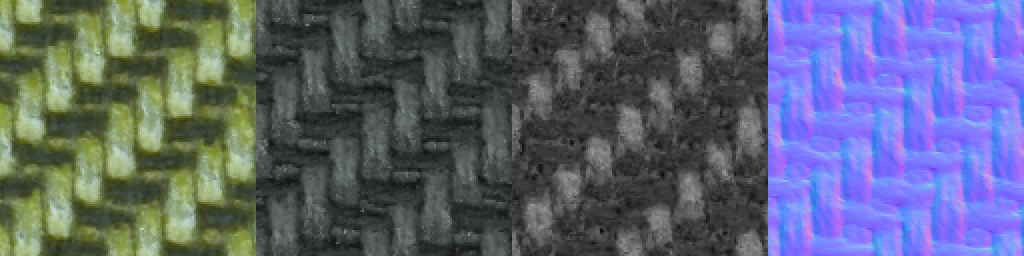} &
    \includegraphics[width=\reswidth]{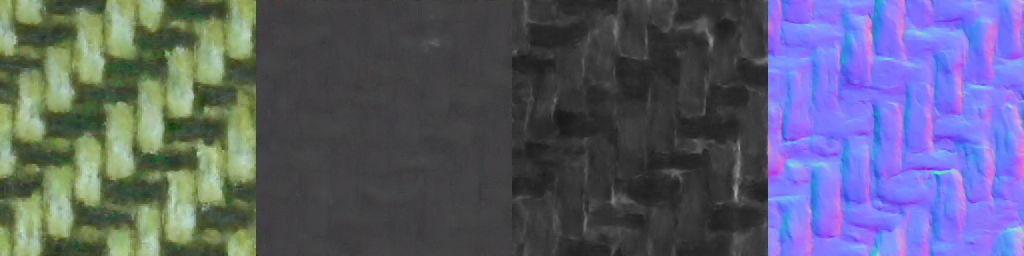} &
    \includegraphics[width=\reswidth]{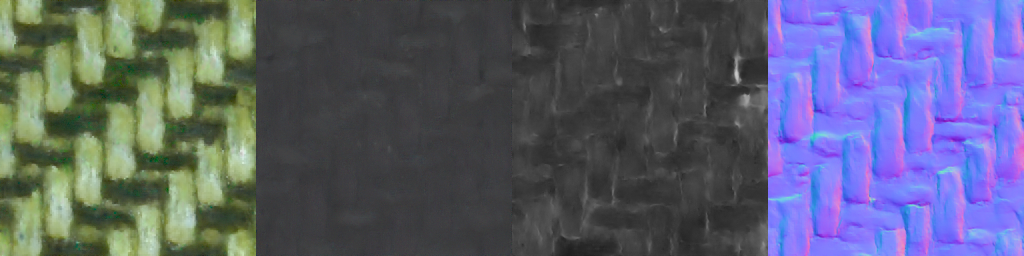} &
    \includegraphics[width=\reswidth]{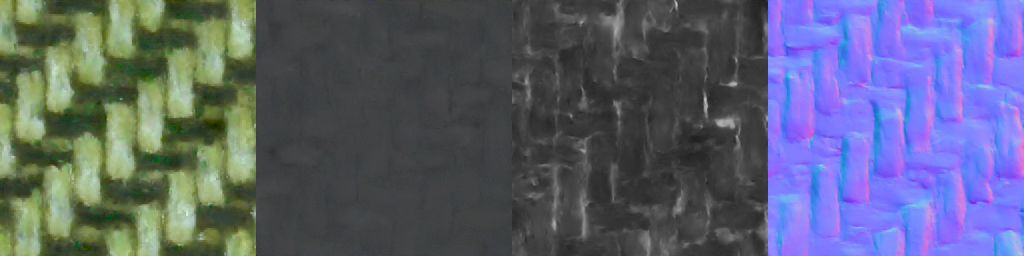} &
    \includegraphics[width=\reswidth]{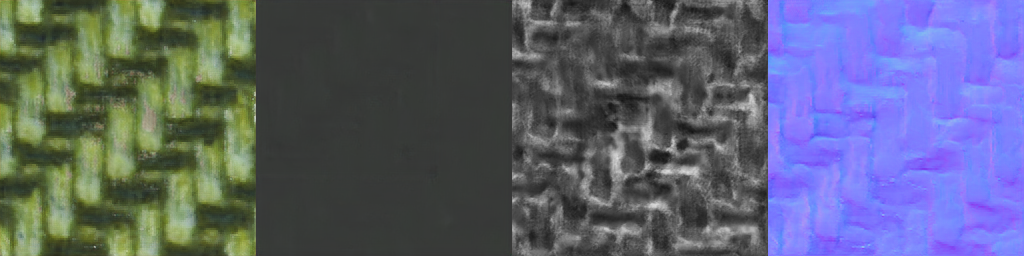} &
    \includegraphics[width=\reswidth]{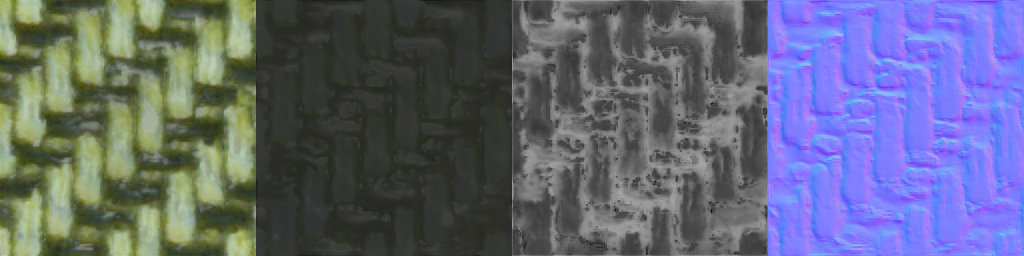} \\

    \multicolumn{2}{r}{Average LPIPS Render Error:} &
    0.3067 &
    0.3129 &
    0.2897 &
    0.3297 &
    0.3213 \\

    \includegraphics[width=\reswidth]{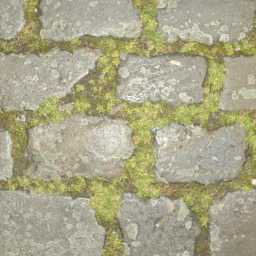} &
    \includegraphics[width=\reswidth]{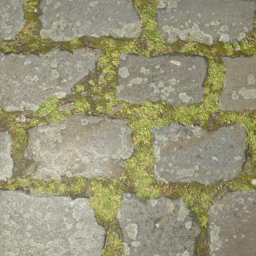} &
    \includegraphics[width=\reswidth]{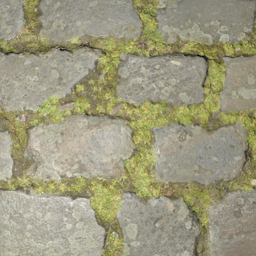} &
    \includegraphics[width=\reswidth]{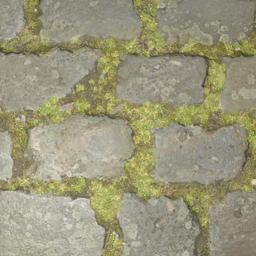} &
    \includegraphics[width=\reswidth]{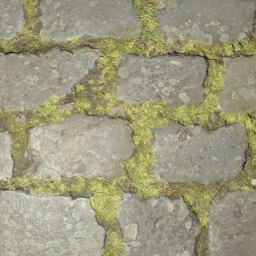} &
    \includegraphics[width=\reswidth]{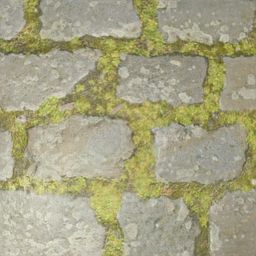} &
    \includegraphics[width=\reswidth]{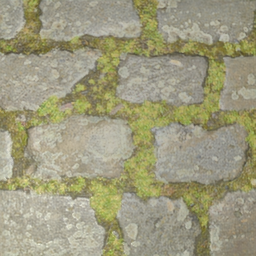} \\

    &
    \includegraphics[width=\reswidth]{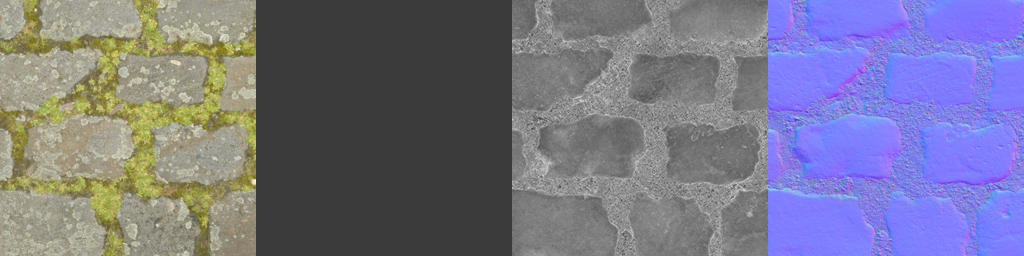} &
    \includegraphics[width=\reswidth]{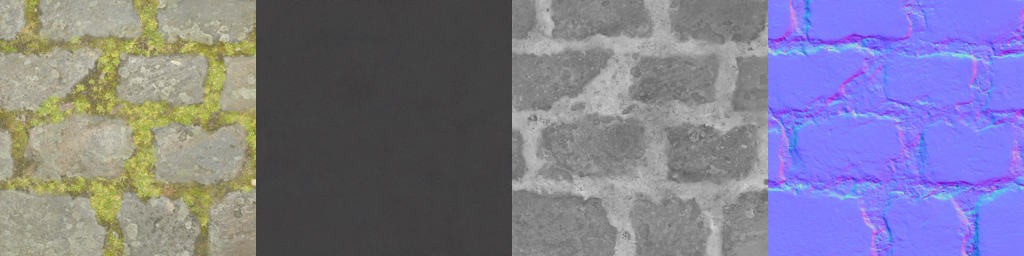} &
    \includegraphics[width=\reswidth]{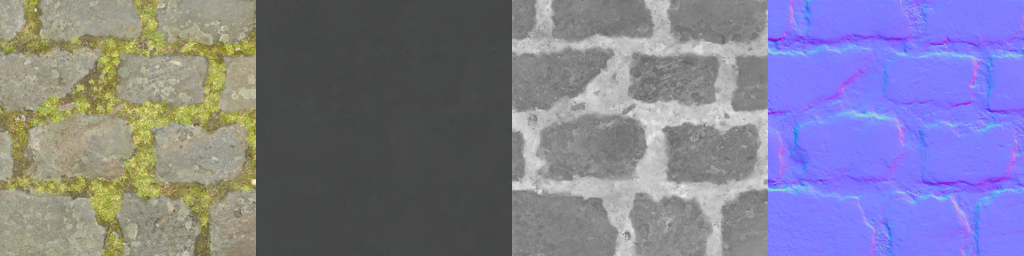} &
    \includegraphics[width=\reswidth]{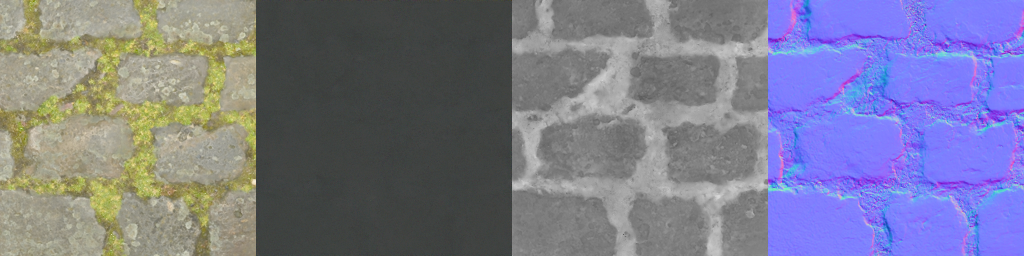} &
    \includegraphics[width=\reswidth]{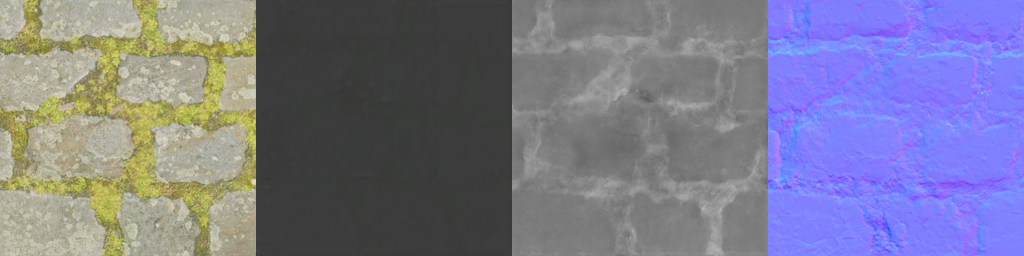} &
    \includegraphics[width=\reswidth]{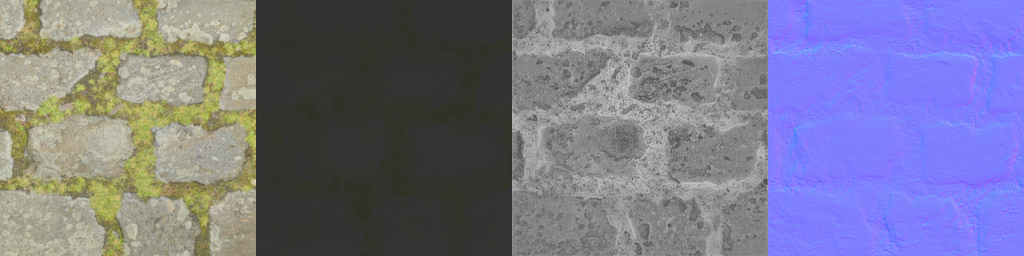} \\

    \multicolumn{2}{r}{Average LPIPS Render Error:} &
    0.2192 &
    0.2317 &
    0.2080 &
    0.2350 &
    0.2656 \\

    \includegraphics[width=\reswidth]{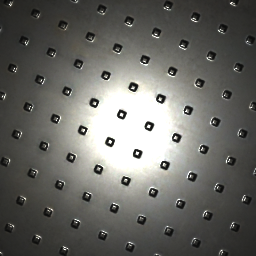} &
    \includegraphics[width=\reswidth]{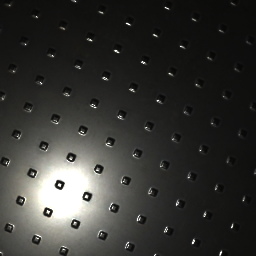} &
    \includegraphics[width=\reswidth]{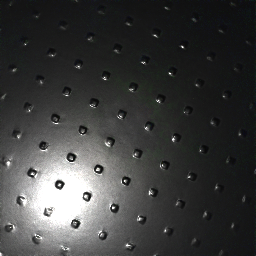} &
    \includegraphics[width=\reswidth]{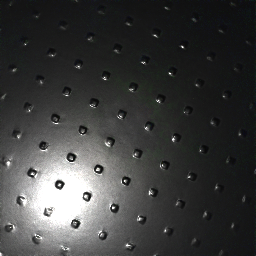} &
    \includegraphics[width=\reswidth]{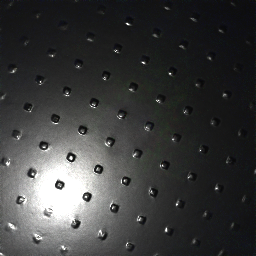} &
    \includegraphics[width=\reswidth]{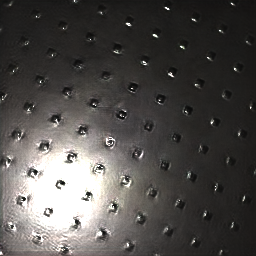} &
    \includegraphics[width=\reswidth]{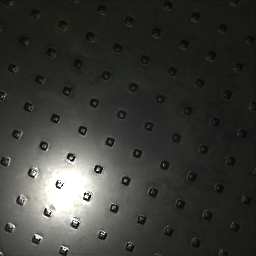} \\

    &
    \includegraphics[width=\reswidth]{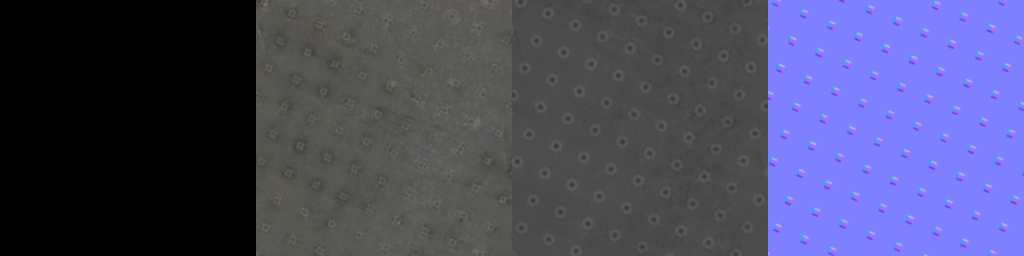} &
    \includegraphics[width=\reswidth]{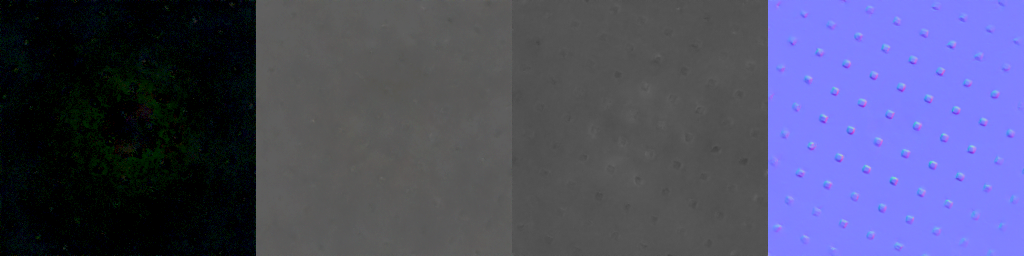} &
    \includegraphics[width=\reswidth]{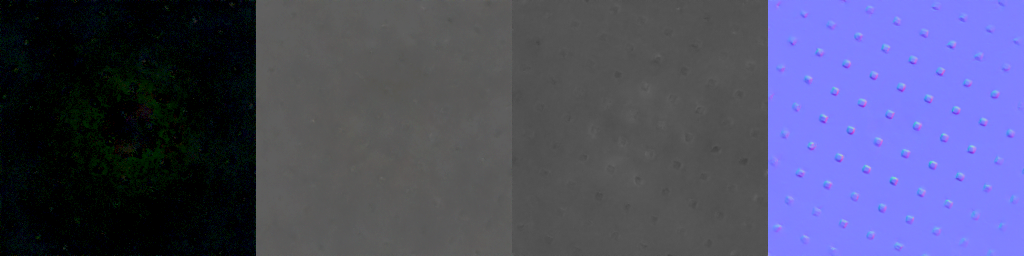} &
    \includegraphics[width=\reswidth]{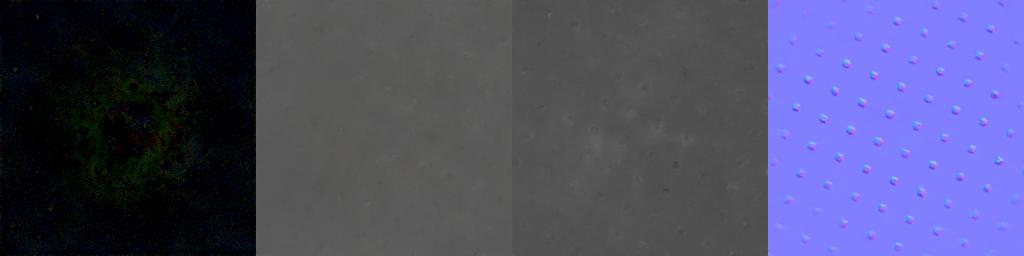} &
    \includegraphics[width=\reswidth]{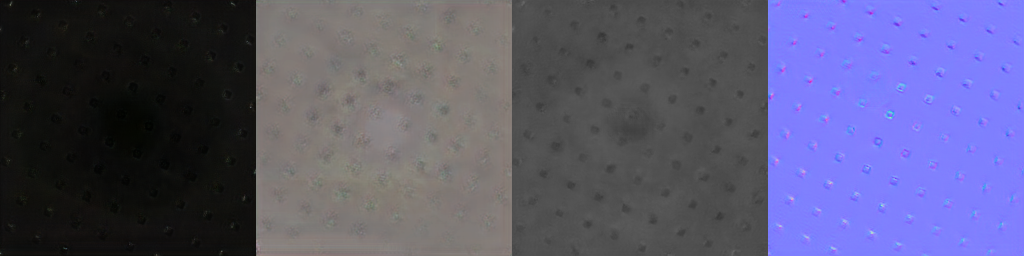} &
    \includegraphics[width=\reswidth]{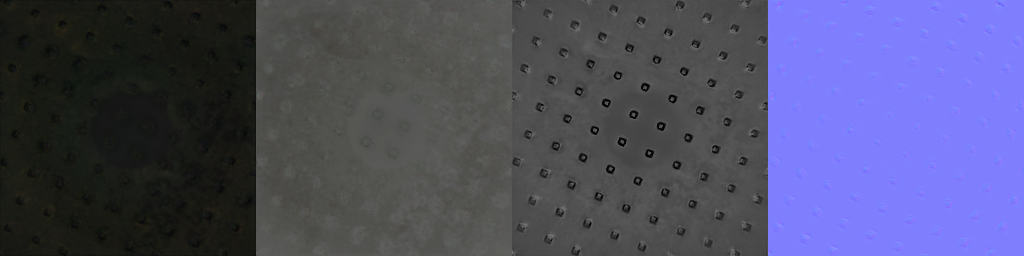} \\

    \multicolumn{2}{r}{Average LPIPS Render Error:} &
    0.1764 &
    0.1764 &
    0.1728 &
    0.2797 &
    0.3967 \\

    \includegraphics[width=\reswidth]{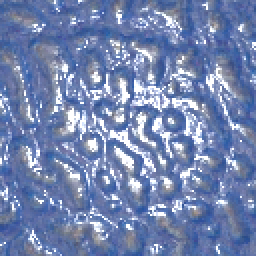} &
    \includegraphics[width=\reswidth]{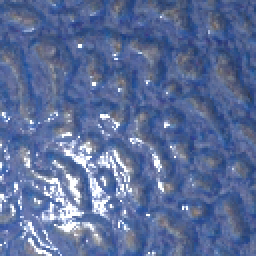} &
    \includegraphics[width=\reswidth]{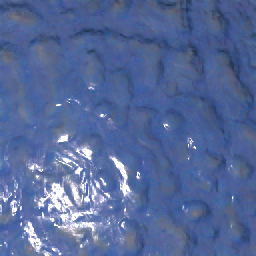} &
    \includegraphics[width=\reswidth]{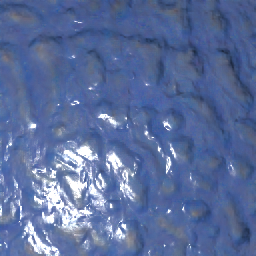} &
    \includegraphics[width=\reswidth]{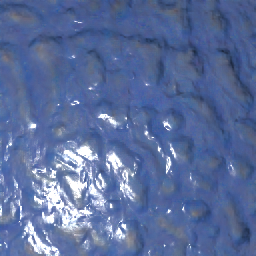} &
    \includegraphics[width=\reswidth]{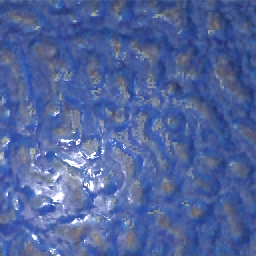} &
    \includegraphics[width=\reswidth]{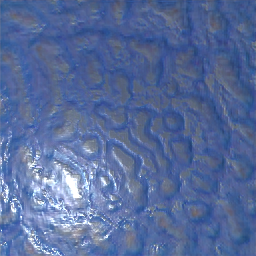} \\

    &
    \includegraphics[width=\reswidth]{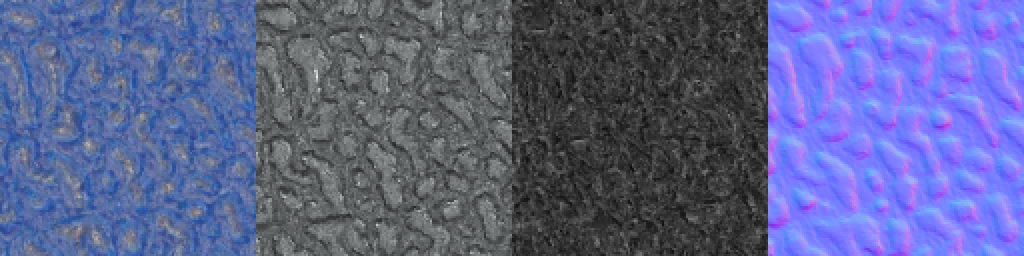} &
    \includegraphics[width=\reswidth]{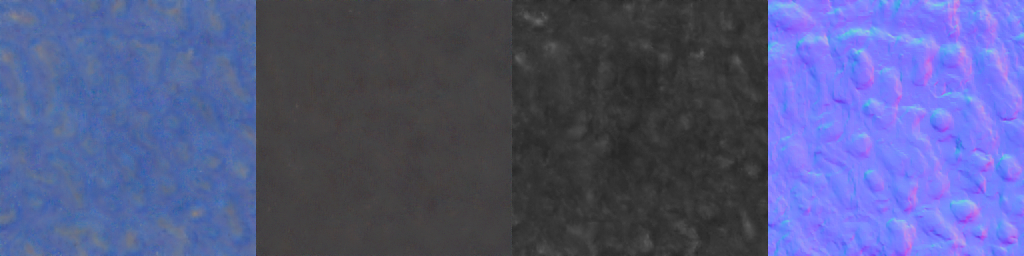} &
    \includegraphics[width=\reswidth]{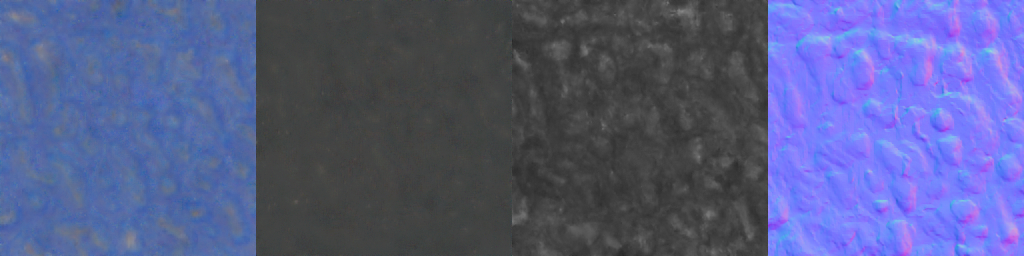} &
    \includegraphics[width=\reswidth]{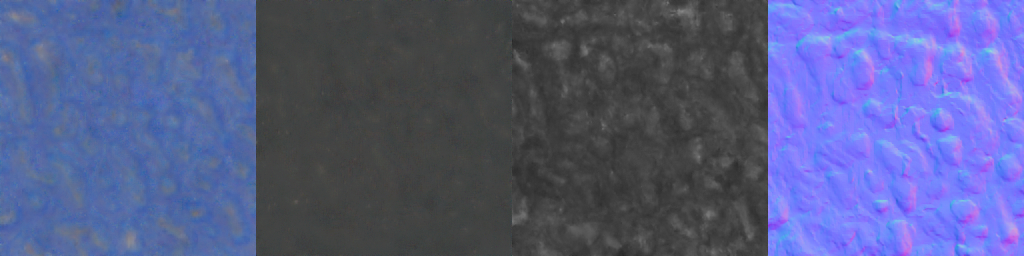} &
    \includegraphics[width=\reswidth]{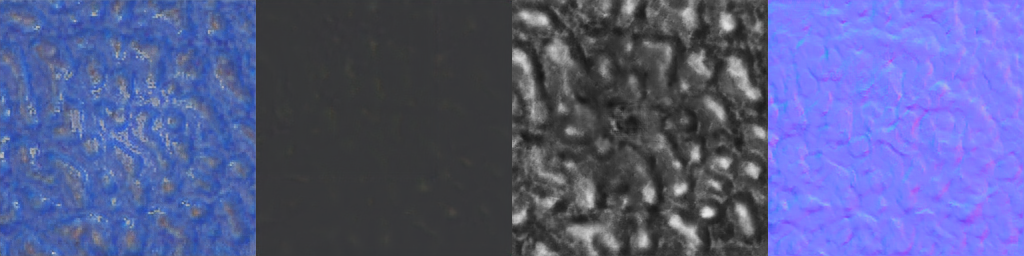} &
    \includegraphics[width=\reswidth]{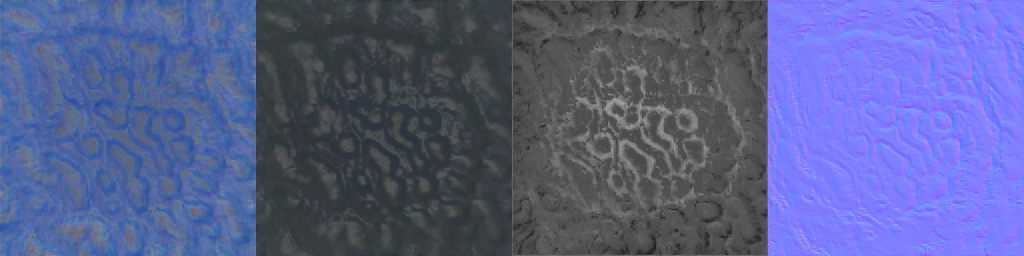} \\

    \multicolumn{2}{r}{Average LPIPS Render Error:} &
    0.3273 &
    0.3044 &
    0.3044 &
    0.3312 &
    0.3932
  \end{tabular}
  }
  \caption{Qualitative comparison of MatFusion conditioned on colocated
    lighting (\emph{fixed seed}, \emph{render error}, and \emph{manual}
    selection) against the adversarial direct inference of Zhou and
    Kalantari~\shortcite{Zhou:2021:ASI} and the meta-leanring look-ahead
    method of Zhou and Kalantari~\shortcite{Zhou:2022:LAT}. The LPIPS errors
    are averaged over visualizations under $128$ different point lights
    sampled on the hemisphere surrounding the sample.}
  \label{fig:comparison}
\end{figure*}

\paragraph{Comparison to Prior Work}
\autoref{fig:comparison} compares the \Variant{colocated} variant for each of
the three selection methods (\emph{fixed seed}, \emph{render error}, and
\emph{manual} selection) against the adversarial direct inference method of
Zhou and Kalantari~\shortcite{Zhou:2021:ASI} and the meta-learning look-ahead
method of Zhou and Kalantari~\shortcite{Zhou:2022:LAT} on synthetic
SVBRDFs. Qualitatively, the \Variant{colocated} model produces a more
plausible appearance and the corresponding property maps appear ``cleaner''.
These qualitative conclusions are supported by the average LPIPS
\cite{Zhang:2018:TUE} render error listed below. We render each exemplar over a
set of $128$ randomly selected point lights on the hemisphere (with a radius of
$2.41$ units to match the training (and thus offer a best case evaluation) of
Zhou and Kalantari~\shortcite{Zhou:2021:ASI,Zhou:2022:LAT}), as well as in
\autoref{tab:relighting} for manual selection on the whole test set of $50$
materials. We argue that a perceptual render error is the best metric for
comparing the different methods as different maps can produce similar material
appearances. For completeness, \autoref{tab:relighting} also lists the RMSE
errors over the SVBRDF property maps. We also include a comparison to Zhou and
Kalantari's adversarial direct inference method retrained using our training
set.  MatFusion is a generative model which does not guarantee pixel-perfect
alignment, which can result in sometimes a larger error on texture-rich
property maps (\eg, 6th row) or unobserved properties (\eg, 2nd row).
However, qualitatively, these property maps include fine details, albeit not
perfectly aligned with the reference.  In contrast, the look ahead-method of
Zhou and Kalantari~\shortcite{Zhou:2022:LAT} produces normal maps with little
detail, resulting in a low error, but distributed over the whole map.
\autoref{fig:comparison} also demonstrates that the render error selection can
provide a good match (\eg, 1st and 5th row), but it can also overfit (\eg, 3rd
row).

\begin{table}
  \def\b{\textbf}
  \begin{center}
    {\small
      \begin{tabular}{r|c c c c c}
                                 & \multicolumn{1}{c|}{LPIPS}  & \multicolumn{4}{c}{RMSE} \\
                                 & \multicolumn{1}{c|}{Render} & Diff.  & Spec.  & Rough. & Normal \\
        \hline
        Adversarial              & 0.2304      & 0.0439      & 0.0859      & 0.1358      & 0.0577      \\
        Adversarial (retrained)  & 0.2292      & \BF{0.0405} & 0.0795      & 0.1276      & 0.0545      \\
        Look-ahead               & 0.2647      & 0.0591      & 0.0727      & 0.1424      & 0.0572      \\
        MatFusion (fixed seed)   & 0.2282      & 0.0427      & 0.0691      & \BF{0.1252} & 0.0561      \\
        MatFusion (render err.)  & \UL{0.2138} & 0.0440      & \BF{0.0657} & 0.1282      & \UL{0.0543} \\
        MatFusion (manual)       & \BF{0.2056} & \UL{0.0412} & \UL{0.0666} & \UL{0.1265} & \BF{0.0524} \\
        \end{tabular}
    }
    \caption{Quantitative comparison of average RMSE on the property
      maps and average LPIPS errors on $128$ renders lit by a uniformly
      sampled point light on the hemisphere for the \Variant{colocated}
      conditioned MatFusion model versus Zhou and
      Kalantari's~\shortcite{Zhou:2021:ASI} adversarial direct inference
      method and Zhou and Kalantari's~\shortcite{Zhou:2022:LAT} meta-learning
      look-ahead method.}
    \label{tab:relighting}
\end{center}
\end{table}

\begin{figure*}[!t]
  \centering
  {
  \small
  \def\reswidthD{0.1\linewidth}
  \def\resdwidthD{0.2\linewidth}
  \def\rsep{\hspace{0.1em}}
  \begin{tabular}{cc@{\rsep}cc@{\rsep}cc@{\rsep}cc@{\rsep}c}
    Input &
    \multicolumn{2}{c}{Reference} &
    \multicolumn{2}{c}{Ours} &
    \multicolumn{2}{c}{\citet{Zhou:2021:ASI}} &
    \multicolumn{2}{c}{\citet{Zhou:2022:LAT}} \\

    \includegraphics[width=\reswidthD]{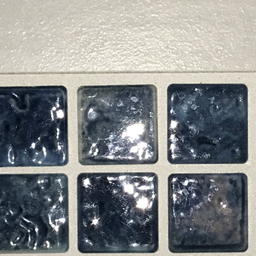} &
    \includegraphics[width=\reswidthD]{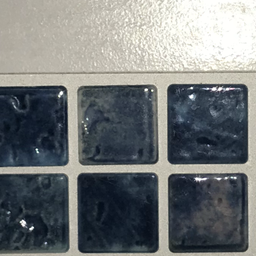} &
    \includegraphics[width=\reswidthD]{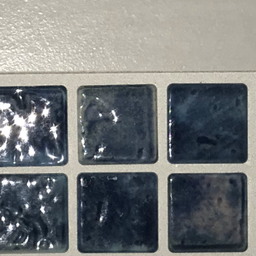} &
    \includegraphics[width=\reswidthD]{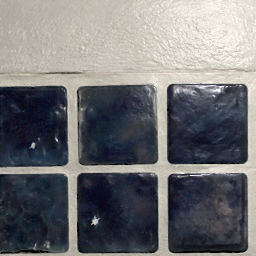} &
    \includegraphics[width=\reswidthD]{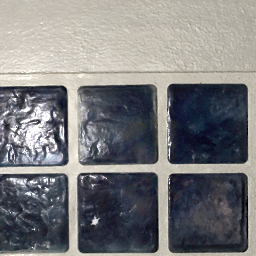} &
    \includegraphics[width=\reswidthD]{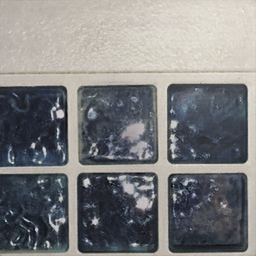} &
    \includegraphics[width=\reswidthD]{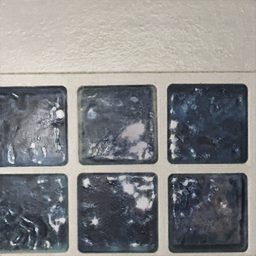} &
    \includegraphics[width=\reswidthD]{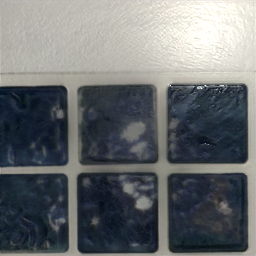} &
    \includegraphics[width=\reswidthD]{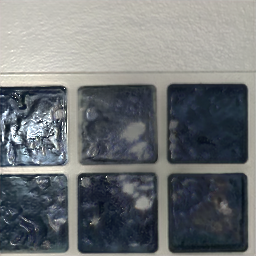} \\

    &&&
    \multicolumn{2}{c}{\includegraphics[width=\resdwidthD]{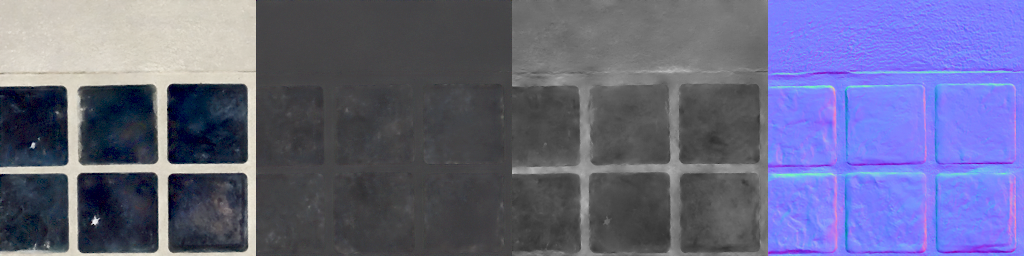}} &
    \multicolumn{2}{c}{\includegraphics[width=\resdwidthD]{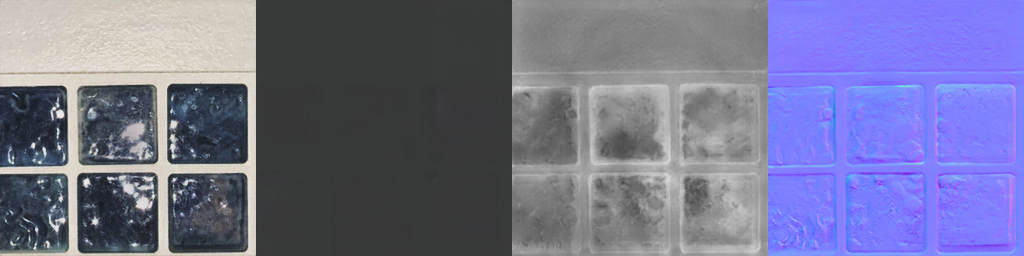}} &
    \multicolumn{2}{c}{\includegraphics[width=\resdwidthD]{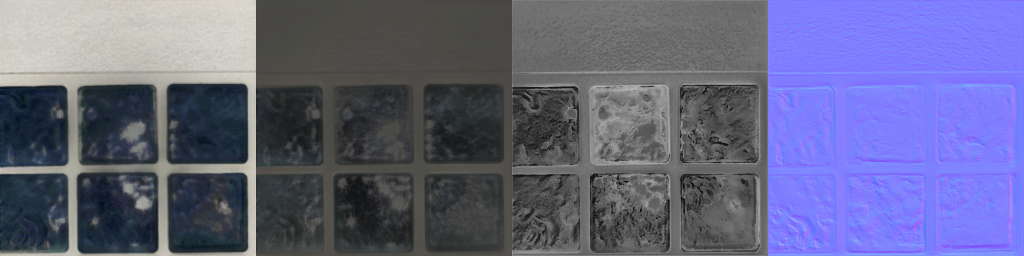}} \\

    \includegraphics[width=\reswidthD]{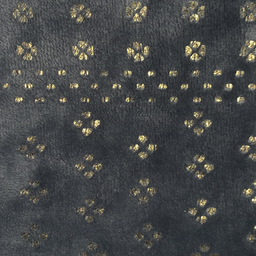} &
    \includegraphics[width=\reswidthD]{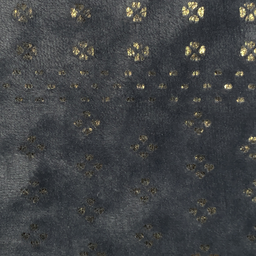} &
    \includegraphics[width=\reswidthD]{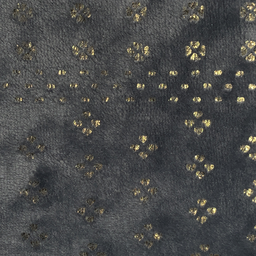} &
    \includegraphics[width=\reswidthD]{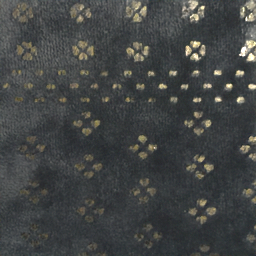} &
    \includegraphics[width=\reswidthD]{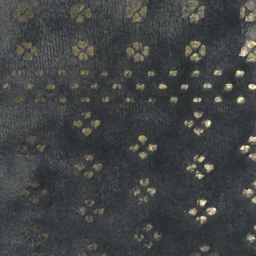} &
    \includegraphics[width=\reswidthD]{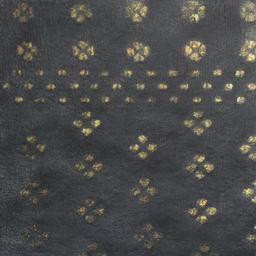} &
    \includegraphics[width=\reswidthD]{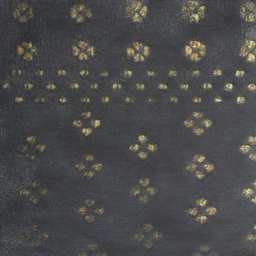} &
    \includegraphics[width=\reswidthD]{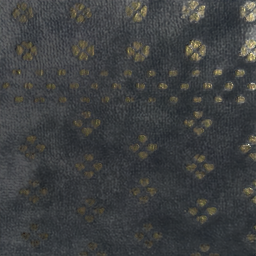} &
    \includegraphics[width=\reswidthD]{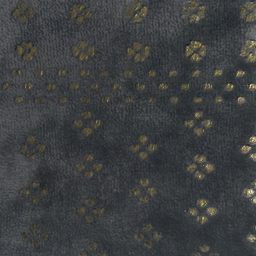} \\

    &&&
    \multicolumn{2}{c}{\includegraphics[width=\resdwidthD]{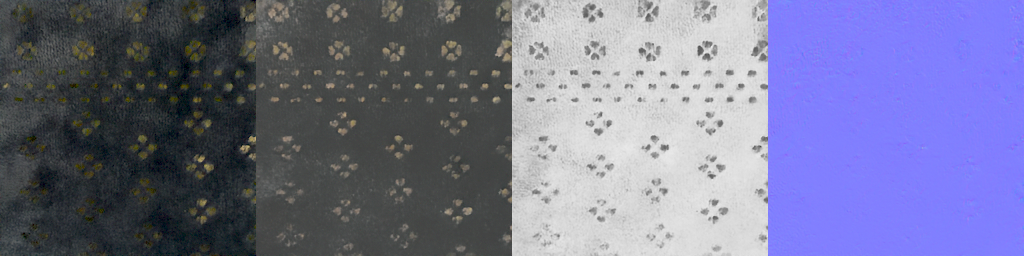}} &
    \multicolumn{2}{c}{\includegraphics[width=\resdwidthD]{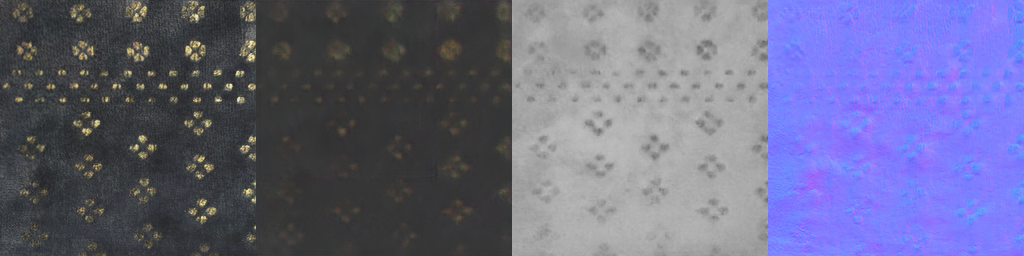}} &
    \multicolumn{2}{c}{\includegraphics[width=\resdwidthD]{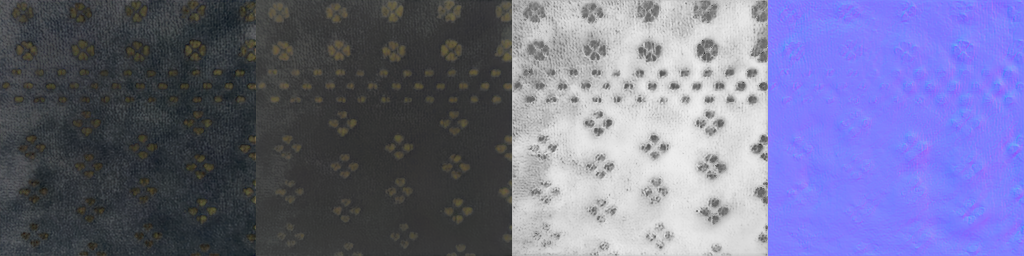}} \\

    \includegraphics[width=\reswidthD]{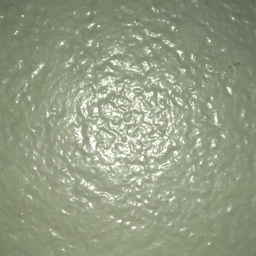} &
    \includegraphics[width=\reswidthD]{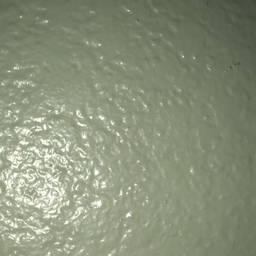} &
    \includegraphics[width=\reswidthD]{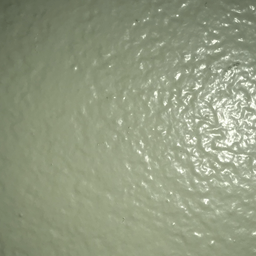} &
    \includegraphics[width=\reswidthD]{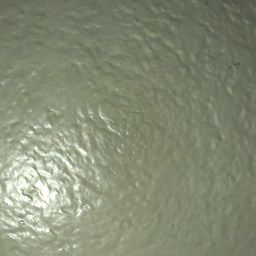} &
    \includegraphics[width=\reswidthD]{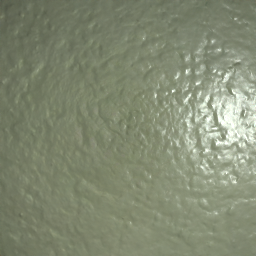} &
    \includegraphics[width=\reswidthD]{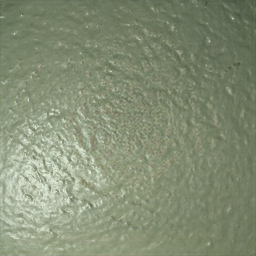} &
    \includegraphics[width=\reswidthD]{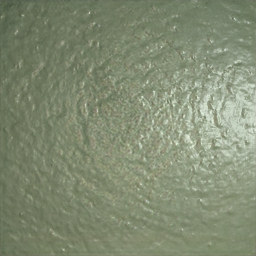} &
    \includegraphics[width=\reswidthD]{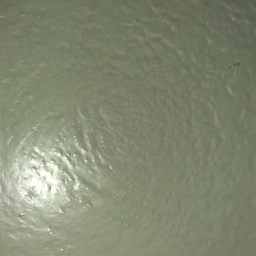} &
    \includegraphics[width=\reswidthD]{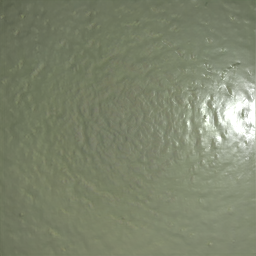} \\

    &&&
    \multicolumn{2}{c}{\includegraphics[width=\resdwidthD]{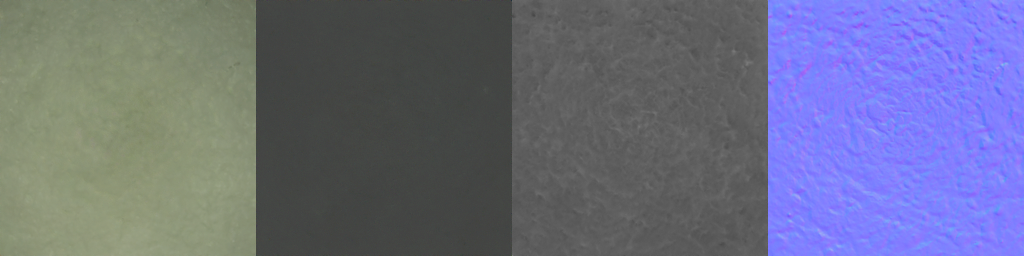}} &
    \multicolumn{2}{c}{\includegraphics[width=\resdwidthD]{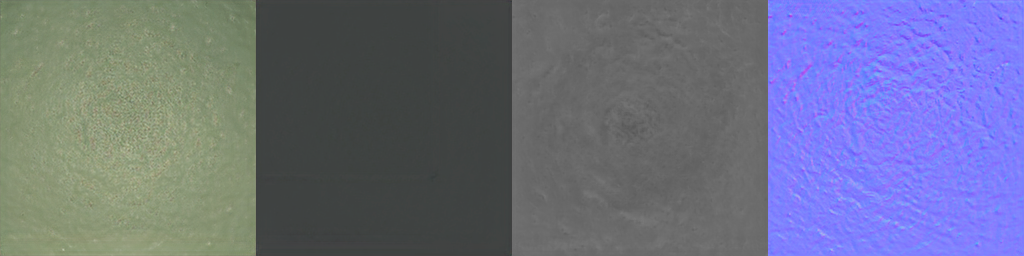}} &
    \multicolumn{2}{c}{\includegraphics[width=\resdwidthD]{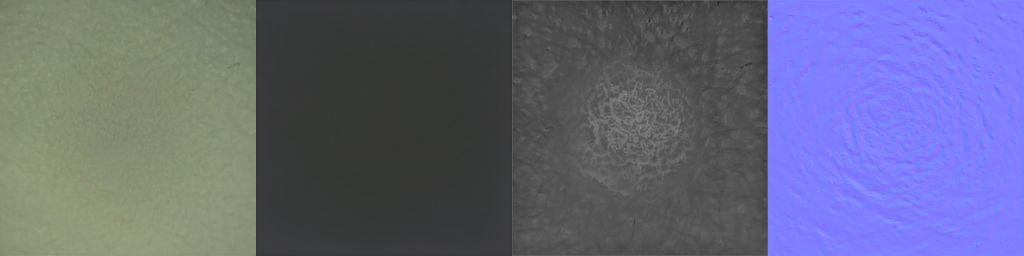}} \\
  \end{tabular}
  }
  \caption{Qualitative comparison on real-world materials captured with a
    colocated light source, and relit from two different point light
    positions.}
  \label{fig:validation}
\end{figure*}

\begin{figure}[!t]
  \centering
  {\small
  \def\reswidthE{0.2\linewidth}
  \def\rsep{\hspace{0.1em}}
  \begin{tabular}{ccc@{\rsep}c}
    Input & SVBRDF & \multicolumn{2}{c}{Render} \\
      \includegraphics[width=\reswidthE]{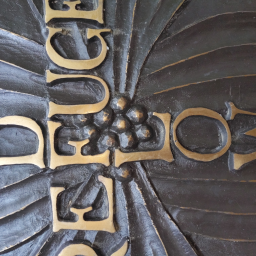} &
      \includegraphics[width=\reswidthE]{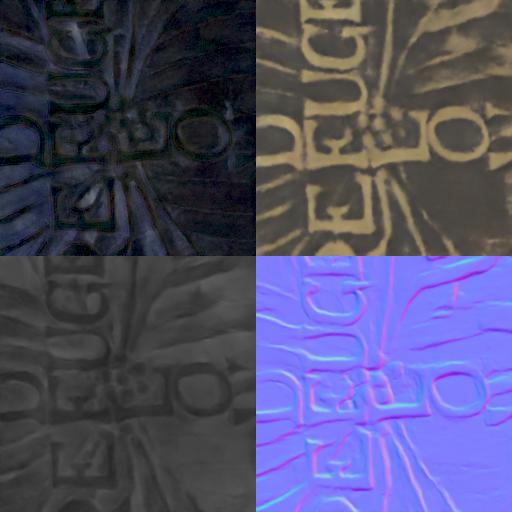} &
      \includegraphics[width=\reswidthE]{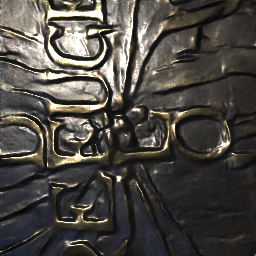} &
      \includegraphics[width=\reswidthE]{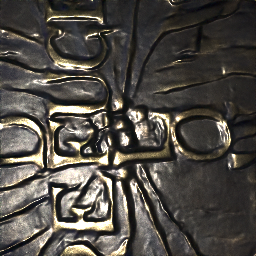} \\
      \includegraphics[width=\reswidthE]{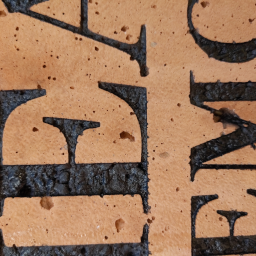} &
      \includegraphics[width=\reswidthE]{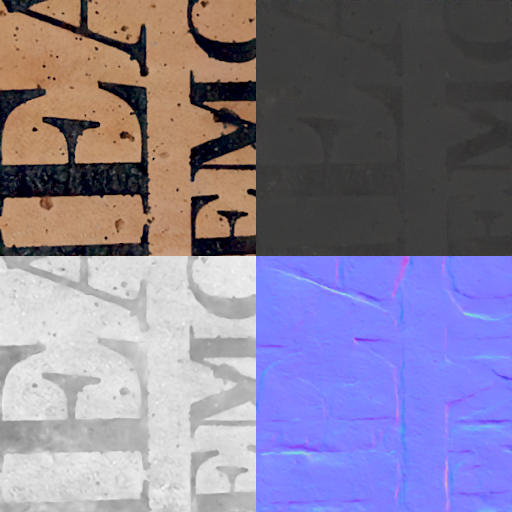} &
      \includegraphics[width=\reswidthE]{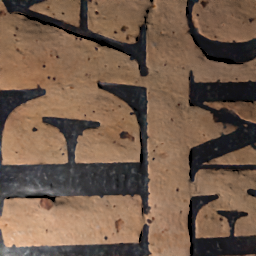} &
      \includegraphics[width=\reswidthE]{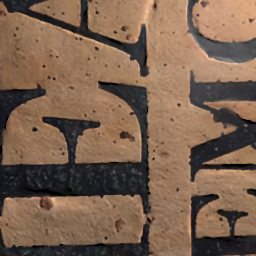} \\
      \includegraphics[width=\reswidthE]{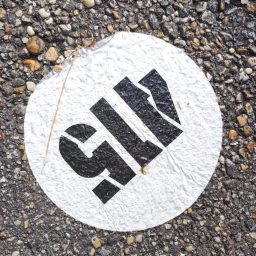} &
      \includegraphics[width=\reswidthE]{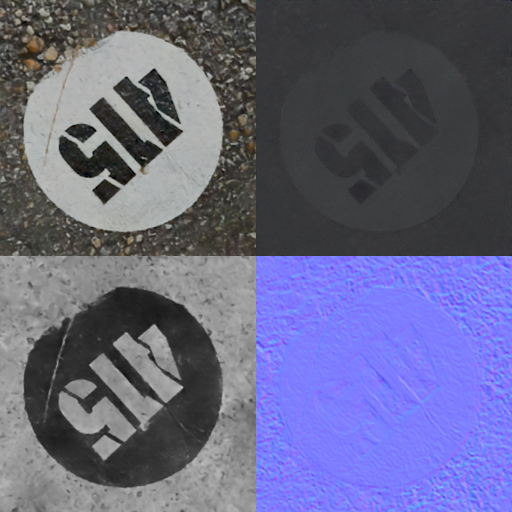} &
      \includegraphics[width=\reswidthE]{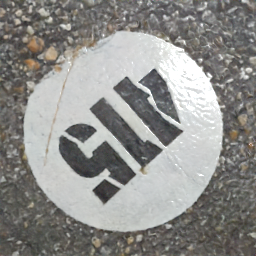} &
      \includegraphics[width=\reswidthE]{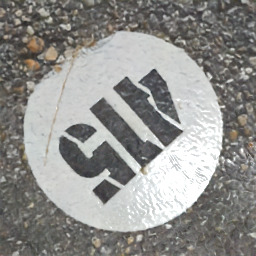}
  \end{tabular}
  }
  \caption{Demonstration of in-the-wild SVBRDF capture under uncontrolled
    unknown natural lighting and revisualized under novel lighting.}
  \label{fig:natural}
\end{figure}

\paragraph{Real-world Validation}
\autoref{fig:validation} and \autoref{fig:natural} demonstrate that MatFusion
generalizes well to real-world captures.  The results
in~\autoref{fig:validation} are manually selected from $10$ random seeds and
validated on the materials captured by Guo~\etal~\shortcite{Guo:2020:MRC}
which also contain reference photographs captured under different
lighting conditions. Our results are visually closer to the reference than the
adversarial direct inference method of Zhou and
Kalantari~\shortcite{Zhou:2021:ASI}, and the look-ahead
method of Zhou and Kalantari~\shortcite{Zhou:2022:LAT}.  Our method suffers
less from specular burn-in (1st example) and overfitting normal detail
to specular highlights in the input (2nd and 3rd example).

The materials in \autoref{fig:natural} are captured in-the-wild by us using a
\emph{Pixel 5a} cell phone, and we manually select the most plausible SVBRDFs.
Note that these images are captured under unknown natural lighting, and due to
the uncontrolled nature of the capture conditions, no reference photographs
under different lighting conditions are available. Nevertheless, the SVBRDF
property maps nicely separate diffuse and specular, and the renderings
plausibly capture the appearance from the input photographs.

\begin{table}
  \begin{center}
    {\small
      \begin{tabular}{r |c c c c c}
        & \multicolumn{1}{c|}{LPIPS}  & \multicolumn{4}{c}{RMSE} \\
        & \multicolumn{1}{c|}{Render} & Diff.  & Spec.  & Rough. & Normal \\
      \hline
      ResNet+Control   & 0.2655 & 0.0525 & 0.0813 & 0.1536 & 0.0545 \\
      ConvNeXt+Control & 0.2731 & 0.0517 & 0.0764 & 0.1428 & 0.0604 \\
      ResNet+Direct    & \UL{0.2093} & \UL{0.0432} & \UL{0.0682} & \BF{0.1055} & \UL{0.0528} \\
      ConvNeXt+Direct  & \BF{0.2056} & \BF{0.0412} & \BF{0.0666} & \UL{0.1265} & \BF{0.0524} 
    \end{tabular}
  }
  \caption{Achitecture ablation study of average RMSE on the property
    maps and average LPIPS render errors on $128$ visualizations lit by a
    uniformly sampled point light, comparing the impact of using Residual
    convolution blocks versus ConvNeXt convolution blocks, and comparing the
    difference between using ControlNet and our direct conditioning.}
    \label{tab:ablation}
\end{center}
\end{table}

\paragraph{Ablation Study}
We perform an ablation study to justify the design decisions with respect to
the architecture of MatFusion (\autoref{tab:ablation}). We validate both the
impact of using Residual versus \mbox{ConvNeXt} convolutional blocks and using
ControlNet versus direct conditioning. For all models we compute the average
RMSE on the property maps and average LPIPS error on renders under the same
set of random point lights for each of the $50$ test materials.
From~\autoref{tab:ablation}, we observe that \mbox{ConvNeXt} layers slightly
outperform Residual convolutional blocks on LPIPS error and $\sim\!5\%$ better
on RMSE on the albedos; the lower roughness error for ResNet is due to a few
outlier materials. Furthermore, direct conditioning outperforms ControlNet on
all metrics, while training time is similar for both, except that ControlNet
requires significantly more memory resources.  We posit that the difference in
performance is due to ControlNet only receiving indirect feedback (by copying
the initial weights) of the diffusion network it aims to control, whereas
direct conditioning closely intertwines both control and synthesis.
Furthermore, our input conditions are more strict, leaving less room for
synthesis than typical ControlNet conditions (\eg, sketches).  However, our
conclusions with respect to ControlNet are only validated for MatFusion using
photographs as conditions, and further investigations are needed to ascertain
whether these conclusions extend to other diffusion networks and/or condition
types.

\begin{figure}[!t]
  \centering
  {\small
    \def\reswidthF{0.2\linewidth}
    \def\rsep{\hspace{0.1em}}
  \begin{tabular}{ccc@{\rsep}c}
    Input & SVBRDF & \multicolumn{2}{c}{Render} \\
      \includegraphics[width=\reswidthF]{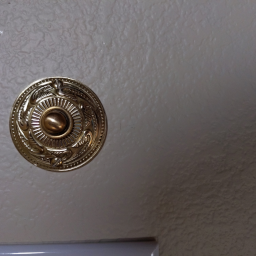} &
      \includegraphics[width=\reswidthF]{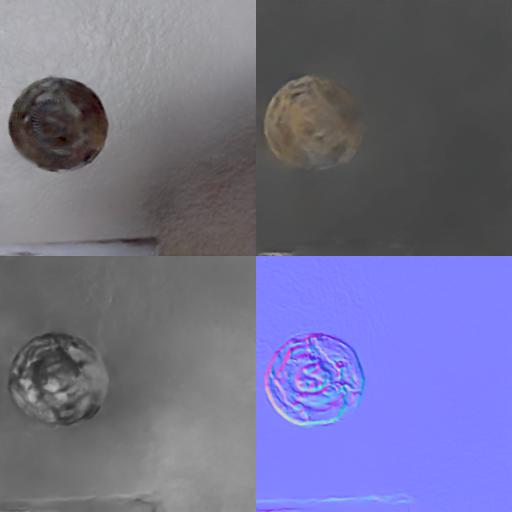} &
      \includegraphics[width=\reswidthF]{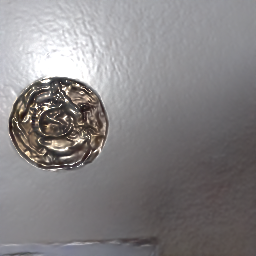} &
      \includegraphics[width=\reswidthF]{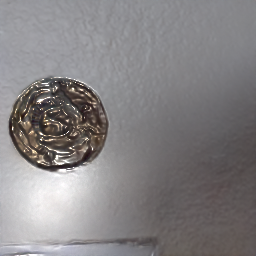} 
  \end{tabular}
  }
  \caption{Failure case: artificial ``blob-like'' normal maps.}
  \label{fig:failure}
\end{figure}

\paragraph{Limitations}
MatFusion is a generative SVBRDF model, and it has trouble generating
pixel-perfect reproductions. Hence, MatFusion does not necessary produce the
lowest errors on pixel-based metrics.  Furthermore, as a generative model,
MatFusion is better suited for capturing materials with organic structures
than those with regular straight lines.  We posit that this is the reason why
MatFusion tends to produce higher quality results on real-world captures than
on artist-generated materials which are more regular.  This causes MatFusion
to sometimes generate properties maps that look too artificial
(\autoref{fig:failure}).  Furthermore, MatFusion is currently limited to
$256 \times 256$ resolution SVBRDFs.  Finally, the render error selection
requires prior knowledge of the lighting condition, hampering automatic
selection from photographs under unknown lighting (\eg, natural lighting).
Furthermore, it does not always yield a good selection because oversaturation
can make it difficult to differentiate between two SVBRDFs that produce a
similar rendered replica but that substantially differ in quality.  Ideally,
we would like to employ a selection criterion that judges plausibility of the
SVBRDFs.

\section{Conclusion}
\label{sec:conclusion}
We presented MatFusion, a generative SVBRDF diffusion model trained on a new
large and diverse training set of synthetic SVBRDFs. MatFusion can
subsequently serve as a starting point for refining an SVBRDF diffusion model
conditioned on captured images under some target lighting condition.  We
demonstrated the flexibility and efficacy of MatFusion by training three
conditional variants: one for photographs captured with a colocated flash
light, one under unknown and uncontrolled natural lighting, and one for
flash/no-flash image pairs.  An advantage of using a generative SVBRDF model
is that different replicates can be synthesized by changing the seed, allowing
user to select the most plausible replicate.  For future work we would like to
investigate more comprehensive metrics for automatic selection, and better
regularization during training and/or inference for modeling regular features.
Based on the recent successes in coupling large language models with diffusion
models, another interesting avenue would be to explore better authoring tools
for SVBRDF creation.

\begin{acks}
This research was supported in part by NSF grant IIS-1909028.
\end{acks}

\bibliographystyle{ACM-Reference-Format}
\bibliography{references}

\end{document}